\pdfoutput=1

\documentclass[11pt,table]{article}

\usepackage[final]{acl}

\usepackage{times}
\usepackage{latexsym}

\usepackage[T1]{fontenc}
\usepackage{float}
\usepackage{soul}

\usepackage[utf8]{inputenc}

\usepackage{microtype}
\usepackage{enumitem}

\usepackage{multirow}
\usepackage{inconsolata}
\usepackage{booktabs}
\usepackage{graphicx}
\usepackage{siunitx} 
\usepackage{array} 

\usepackage{times}
\usepackage{latexsym}

\usepackage[T1]{fontenc}
\usepackage{float}
\usepackage{soul}

\usepackage[utf8]{inputenc}

\usepackage{microtype}
\usepackage{enumitem}

\usepackage{multirow}
\usepackage{inconsolata}
\usepackage{booktabs}
\usepackage{graphicx}
\usepackage{siunitx} 
\usepackage{array} 

\title{My LLM might Mimic AAE - But When Should it?}
\author{Sandra C. Sandoval\thanks{Asterisks (*) indicate similar levels of contribution.} \\ 
        University of Maryland\\
        \texttt{sandracs@umd.edu} \\\And
        Christabel Acquaye\textsuperscript{*}\\
        University of Maryland\\
        \texttt{cacquaye@umd.edu} \\
        \AND
        Kwesi Cobbina\textsuperscript{*} \\ 
        University of Maryland \\
        \texttt{kcobbina@umd.edu} \\\And
        Mohammad Nayeem Teli \\ 
        University of Maryland \\
        \texttt{nayeem@umd.edu} \\\And
        Hal Daum\'{e} III  \\ 
        University of Maryland\\
        \texttt{hal3@umd.edu} \\
        }

\newcommand{\sandra}[1]{\textcolor[rgb]{1.0,0.1,0.4}{[Sandra: {#1}]}}
\newcommand{\chris}[1]{\textcolor[rgb]{1.0,0.5,0}{[Christabel: {#1}]}}
\newcommand{\kac}[1]{\textcolor[rgb]{0.9,0.7,0.1}{[Cobbina: {#1}]}}
\newcommand{\hal}[1]{\textcolor[rgb]{0.7,0.1,0.7}{[{Hal: {#1}}]}}
\newcommand{\nayeem}[1]{\textcolor[rgb]{0.4,0.5,0.6}{[Mohammad: {#1}]}}

\renewcommand{\sectionautorefname}{\S\!\!}
\renewcommand{\subsectionautorefname}{\S\!\!}
\renewcommand{\subsubsectionautorefname}{\S\!\!}

\begin{document}
\maketitle

\captionsetup{font=small}

\begin{abstract}
We examine the representation of African American English (AAE) in large language models (LLMs), exploring (a) the perceptions Black Americans have of how effective these technologies are at producing authentic AAE, and (b) in what contexts Black Americans find this desirable.
Through both a survey of Black Americans ($n=$ 104) and annotation of LLM-produced AAE by Black Americans ($n=$ 228), we find that Black Americans favor choice and autonomy in determining when AAE is appropriate in LLM output. They tend to prefer that LLMs default to communicating in Mainstream U.S. English in formal settings, with greater interest in AAE production in less formal settings. When LLMs were appropriately prompted and provided in context examples, our participants found their outputs to have a level of AAE authenticity on par with transcripts of Black American speech. Select code and data for our project can be found here: \url{https://github.com/smelliecat/AAEMime.git} 


\end{abstract}

\section{Introduction}
\label{sec:intro}
In our study, we explore how an underserved population,  Black Americans in the United States, regards increasingly ubiquitous text-based AI tools in terms of their preferred functionalities and with respect to the authenticity of the language produced by these systems, given their unique needs.
We specifically investigate the research questions:
\begin{enumerate}[nolistsep,noitemsep]
\item \emph{Do Black Americans\footnote{%
We use \textit{Black Americans} to describe those who identify as American with ancestral roots to Black African ethnic groups.}
want generative AI technologies to produce African American English? If so, in what contexts?}; and
\item \emph{How effective are large language models (LLMs) at generating authentic African American English (AAE)\footnote{%
African American English is ``the grammatically patterned variety of English used by many, but not all and not exclusively, African Americans in the United States'' \cite{grieser2022black}; AAE has many alternative names, including African American Language, African American Vernacular of English, Black English, (Black) Slang, and Ebonics
\cite{green2002african,wolfram2015american,rickford2016language,king2020african,becker2013ethnolect}. In our study, participants could choose the terminology they preferred. 
}
when prompted to do so?}
\end{enumerate}

\noindent
The Black American population makes up approximately 13.6\% of the United States total population in 2022 (\citealp{USCensus2022,Moslimani2023}) and represents a major stakeholder population of text-based AI technologies. 
AAE, while predominantly a spoken language variety, is seeing increased representation in speech-like media such as texting and social media \cite{blodgett2016demographic}. 
Its use has become synonymous with the cultural identity of some Black Americans \cite{BashirAli2006} with the language evolving over an extended period of time dating as far back as the period of Black enslavement in the United States. In spite of the cultural importance of AAE, Black Americans have had good reason to be hesitant to use the language outside of personal contexts due to widespread linguistic discrimination: racial identification and discrimination based on speech or writing in the work place and in schools~\cite{baugh2005linguistic}. In addition, the use of AAE may be associated with poverty or lower socioeconomic class \cite{rickford2015neighborhood}, which could influence Black Americans to be cautious about the circumstances under which it should be used.

AAE is predominantly 
identified by grammatical patterns such as the use of double negatives, variable subject-verb agreement, and omission of verbal copulas. These patterns distinguish it sharply from Mainstream U.S. English (MUSE)\footnote{We refer to the most prevalent variety of American English as Mainstream U.S. English  \cite{baker2020dismantling,harris2022exploring}; as with AAE, other names exist, such as Standard American English and White Mainstream English~\cite{wolfram2015american}.}.   
Despite its distinct linguistic characteristics and the large proportion (80-90\%) of the Black American society that speaks AAE in the United States (\citealp{Holt2018, OQuin2021,Farrington2021}), prior studies have shown that AI technologies often fail to accommodate its nuances, especially in the context of speech recognition (\autoref{sec:related_works}).


In this paper, we investigate the research questions listed above in the context of large language models. We conduct an online study (\autoref{sec:methodology}) among Black Americans ($n=$ 104) consisting of both a survey---to understand what is wanted out of this technology---and data annotation---to understand the efficacy of current LLMs at producing authentic AAE.
In the survey (\autoref{subsec:survey}), we aim to understand in what social contexts Black Americans would want (the option of) having an LLM use AAE, including both professional and personal settings, and including both continuation behavior (e.g., email autocomplete) and reply behavior (e.g., AI assistants).
In the data annotation task (\autoref{subsec:data_annotations}), we ask Black Americans ($n=$ 228 who provided 8,654 judgments for 1,357 examples) to judge text generated by three different LLMs---GPT 4o-mini, Llama 3, and Mixtral---along axes like coherence, the explicit presence of AAE features, and offensiveness.
Our main contributions (\autoref{sec:results}) include:
\begin{enumerate}[nolistsep,noitemsep]
\item We find that Black Americans favor the use of MUSE in more formal or task-specific interactions, but are open to LLM generation of AAE in personal or casual settings, preferring the autonomy to switch to AAE as desired.
\item We find that Black Americans judged the LLM AAE generations as equally authentic to the human baseline (Black American transcribed interviews), and they did not consider them to be mocking or offensive.
\item We contribute a dataset of linguistic judgments from Black American annotators on both AAE and MUSE texts, drawn both from human- and LLM-produced text. In addition, we share the dataset and a selection of our code for the project here: \url{https://github.com/smelliecat/AAEMime.git}
\end{enumerate}




\section{Related Work} \label{sec:related_works}

\subsection*{Attitudes and Perceptions of AAE Speakers Towards Technology} 
\citet{Cunningham2024Impacts} examine the invisible labor AAE speakers undertake to be understood by language technologies. Their findings show that AAE speakers often have to proactively adapt their speech, which leads to significant frustration and alienation when interacting with these systems.
\citet{harrington} further discuss how Black older adults anticipate and experience substantial challenges with voice assistants, exacerbating their reluctance to use these devices for tasks like health information seeking. Our study explores the idea that Black Americans may prefer AAE options in their interactions with speech or language-based AI technologies. We build on these lines of research by exploring Black Americans' preferences for AAE representation in specific contextual settings.



\subsection*{Current State of AAE in Technology}

\citet{hill1998language} and \citet{smokoski2016voicing} highlight the issue of AAE in social media as not always true AAE, but rather non-Black Americans mimicking or mocking AAE. Thus, one of our objectives for this study was to begin to understand some of the constraints that generative AAE must adhere to, to stay within the bounds of acceptability to the AAE community, such that the language is not seen as mocking individuals who use the language and remains relatable and respectful to their specific use context. Consequently, we address the idea of whether the AAE generated by LLMs could be construed as mocking or offensive in our study.


With respect to the importance of language diversity being reflected in chat-based AI technologies, \citet{Harrington2019DeconstructingCC} critique how participatory design often overlooks historical inequities, further marginalizing AAE speakers. This reflects broader systemic issues in technology design that fail to accommodate linguistic diversity. 

\citet{santiago2022disambiguation} highlight the critical role of morphosyntactic features in AAE, such as the invariant `be', demonstrating how leveraging these features improves the disambiguation of syntactic constructs, and mitigates the risks of discrimination in Automatic Speech Recognition (ASR) systems when misrepresented. In the context of large language models, \citet{HYFIN}'s `Black GPT' is an example of an AI technology that has been developed with recognition of the value of AAE, and thereby promotes inclusion and representation of Black Americans 
\cite{previlon-etal-2024-leveraging-syntactic}.\looseness=-1

Despite recent efforts to provide inclusive technologies that meet the needs of Black Americans, \citet{pinhanez2024creating} highlight the ongoing challenges to developing systems that authentically represent AAE. In particular, they look at text-to-speech systems, uncovering latent biases that prevent broad recognition and acceptance without resorting to stereotypes. Issues in ASR have been documented for systems from Apple, Amazon, Google, and IBM, which exhibit error rates for Black speakers of 35\%, nearly double those for White speakers at 19\% \cite{Koenecke2020}.

This discrepancy highlights a systemic oversight in AI design that fails to consider AAE’s unique linguistic features, necessitating that AAE speakers frequently engage in code-switching. This adaptation requires significant invisible labor and can lead to alienation and frustration, as illustrated by Harrington's examination of Black older adults using voice assistants \cite{harrington} and Cunningham's insights into the emotional toll of such adjustments \cite{Cunningham2024Impacts}.

Unfortunately, the impacts of design oversights extend to chatbots and LLMs as well. \citet{hofmann2024dialect} expose how LLMs covertly perpetuate dialect prejudice, with their Matched Guise Probing approach, highlighting that even when trained with human feedback, these systems still enforce negative stereotypes, impacting judgments on employability and criminality.

In our study, we strive to uncover if LLMs are able to generate AAE effectively as judged by Black Americans themselves, and if they suffer from some of the same biases as speech recognition AI technologies according to our study participants. We note the distinction of the large language model-based chatbots we study here as being tasked to generate AAE, in contrast to speech recognition technologies which should understand AAE.


\section{Methodology} \label{sec:methodology}




We investigate two primary research questions: 1) whether Black Americans want Generative AI technologies to communicate and understand African American English and in which contexts, and 2) how effective LLMs are at generating AAE when prompted to do so. Our Institutional Review Board (IRB)-approved\footnote{Our IRB restricts us from reporting on our survey participant and annotator demographics other than at the aggregate level. Most other IRB details redacted for anonymity.} online study gathers participant feedback on these questions, via a survey on participants' desires (\autoref{subsec:survey}), and data annotation to understand the effectiveness of current LLMs (\autoref{subsec:data_annotations}). 

For both, we recruit participants through Prolific (\url{prolific.co}), who were required to be Black American over 18 years old, familiar with AAE, and reside in the U.S. 
In total, of our $n=104$ \textit{survey} participants, $61$ identified as female, $37$ as male, $3$ as non-binary, $1$ as other (with $2$ undisclosed). Participants ranged from $25$ to $64$ years old ($\mu=34$), with the majority (81 participants) 
having attended at least some college.
The participants come from all four major geographic areas in the U.S. ($53$ from the South, $21$ Northeast, $17$ West, and $17$ Midwest) (see \autoref{sec:annotators}). For the \textit{data annotation} tasks, we recruited 
$n=228$ 
Black American annotators with similar demographics.

\subsection{Survey of If and When Black Americans Want LLMs to Produce AAE} \label{subsec:survey}
Our survey is designed to explore the perceptions and attitudes of Black Americans regarding AAE representation in chat-based AI systems across a variety of settings, ranging from professional to personal. For each setting, we gauge how and when participants want an LLM (or chatbot) to use AAE versus MUSE. We aimed to provide sufficient detail on the settings to make them more relatable and easier to comprehend \cite{lenzner2012effects}. The settings are selected to give a more complete picture of Black American's every-day experiences and preferences \cite{Maedche2019AI-Based}. For each scenario presented, study participants were asked to choose from the answer choices seen in \autoref{table:Vignette Choices} for how they would want such an LLM to interact:



\begin{enumerate}[nolistsep,noitemsep]
\item \textit{AI Assistants} (professional and personal).
These LLM-response settings reflect the use of an AI assistant for helping with either professional or personal tasks, and whether such an assistant should address the user in AAE.

\item\textit{Customer Bot}. This LLM-response setting reflects the use of a text-based chatbot agent for quick assistance, and whether the agent should continue the interaction after greeting the user in AAE.

\item \textit{Email and SMS Autocomplete}. These LLM-continuation settings reflect the use of an LLM to autocomplete a user's own writing for emails or text messages.  

\item \textit{Educational Avatar}. This LLM-response setting reflects the use of AAE by an avatar in an education platform and whether this could impact learning experience.
\end{enumerate}


\noindent



\begin{table}[t]
\centering
\footnotesize
\renewcommand{\arraystretch}{0.85} 
\setlength{\tabcolsep}{4pt}
\begin{tabular}{ @{~}l @{~~~} p{55mm} @{~} }
  \toprule			
AlwaysMUSE &	LLM should always use MUSE.	\\
AlwaysAAE &	LLM should always use AAE. \\[0.5em]
AutoDetect &	LLM should automatically detect/adapt to\\&\quad the user's language variety. \\
UserOption &	LLM should provide an option to switch\\&\quad between AAE and MUSE. \\[0.5em]
NoPreference	&	No preference as long as the system is\\&\quad effective. \\
  \bottomrule			
\end{tabular}	
\caption{The set of possible choices for the preferences survey, which asked Black Americans about the contexts or scenarios in which they would prefer to have language model-based AI technologies generate AAE vs MUSE.} 
\label{table:Vignette Choices}
\end{table}

\subsection{Annotation of LLM Output for AAE-ness} \label{subsec:data_annotations}



\begin{table}[t]
\centering
\footnotesize
\renewcommand{\arraystretch}{0.85} 
\setlength{\tabcolsep}{4pt}
\begin{tabular}{ @{~}l @{~~~} p{55mm}@{~}}
  \toprule	
  \textbf{Description} & \textbf{Linguistic Judgment} \\
  \midrule			
\textit{Coherent}	&	The text is a coherent
continuation.	\\
\textit{AAE Features}	&	The text contains AAE features. \\
\textit{Black Sounding}	&	The text sounds like something a Black \\&\quad American would say.\\
\textit{White Sounding}	&	The text sounds like something
a White\\&\quad American would say.\\
\textit{Mocking}	&	The text is like someone mocking AAE.\\
\textit{Offensive}	& I would be offended if a chatbot said this. \\
  \bottomrule			
\end{tabular}	
\caption{Assessments made via Likert score rating by Black American annotators regarding the AAE and MUSE text continuations they reviewed. The continuations were either human or LLM-produced, but annotators were not told which.} 
\label{table:Label Descriptions}
\end{table}



We used text annotation to explore our second research question: how well do current LLMs generate AAE-like text (including relative to generation of MUSE-like text). We obtained human judgments by our Black American annotators as to how they perceived LLM-produced AAE text relative to their expectations for AAE. 
We show participants text, including a prefix which was transcribed (human) speech from interviews or from X (Twitter) posts, paired with a suffix which was either the natural (human) continuation or an LLM-generated continuation (by GPT, Llama, or Mixtral; see \autoref{fig:linguistic_examples} for examples).
In some cases the human (and LLM-generated) text is AAE; in others it is MUSE.
We adopt this \textit{continuation} methodology to ensure that that LLM generations are comparable---similar topic, etc.---to the human generations to facilitate more controlled comparisons.
In the annotation task, the suffix was highlighted and participants were asked to assess it along the six dimensions from \autoref{table:Label Descriptions} on a five-point Likert scale from ``Strongly Disagree'' to ``Strongly Agree''
(the full interface annotators saw is shown in \autoref{fig:annotation_task} in the Appendix, and see \autoref{fig:linguistic_examples} for examples with annotations).
Participants were unaware if the suffix was human or machine-generated. 

 The AAE and MUSE prefixes and human baseline texts for this annotation task were drawn from the Corpus of Regional African American Language (CORAAL) \cite{KendallFarrington2023}, which is a corpus of transcribed interviews of Black Americans, an X posts (Twitter) AAE corpus \cite{aave_corpora}, and a National Public Radio interview corpus containing MUSE text \cite{majumder2020large,majumder2020open} 
 Hereafter, we refer to X posts as Tweets for brevity.
 We lightly cleaned the corpora for annotator assessments (see \autoref{apdx:corral_desc}). 

\subsection{Prompting LLMs}

To produce the LLM suffixes, for each prompt, we provided the following inputs to the LLMs to encourage the systems to generate completions of the interviewee statements in AAE for the CORAAL interview and X post (Twitter) prefixes, or in MUSE for the NPR interview-sourced prefixes: 1) an instruction to generate a completion (the suffix) \textit{in AAE or in MUSE} with a suggested response length limit; 2) a list of 2-5 randomly sampled (and randomly varying in number of) training examples of authentic AAE (or MUSE for those continuations) for in context learning~\cite{brown2020language} -- the AAE training examples were drawn from other CORAAL interview exchanges or other AAE Tweets, including both the interviewer statements and the corresponding interviewee responses, or the original Tweets; and 3) a test prefix as described above, that the LLM then was tasked to continue in AAE (or MUSE for the texts originally from NPR interviews). See \autoref{apdx:gtp} for details and \autoref{apdx:prompts} for examples of the prompt templates used.


\section{Results} \label{sec:results}


We outline the results from our study below. We review our findings from our scenarios-based survey, which explored Black American perspectives and expectations for interactions with language-based generative AI tools, looking at a range of personal and professional settings. We then review the results from the data annotation effort, whereby we gathered Black Americans' judgments of the effectiveness of large language models at production of AAE and MUSE relative to our human baselines of Black American transcribed interviews -- CORAAL, Tweets, and NPR Interviews.





\subsection{Survey (Scenarios-based Questions)}

\begin{figure}[t]
\centering
\includegraphics[width=0.85\columnwidth,trim=8 32 60 8,clip]{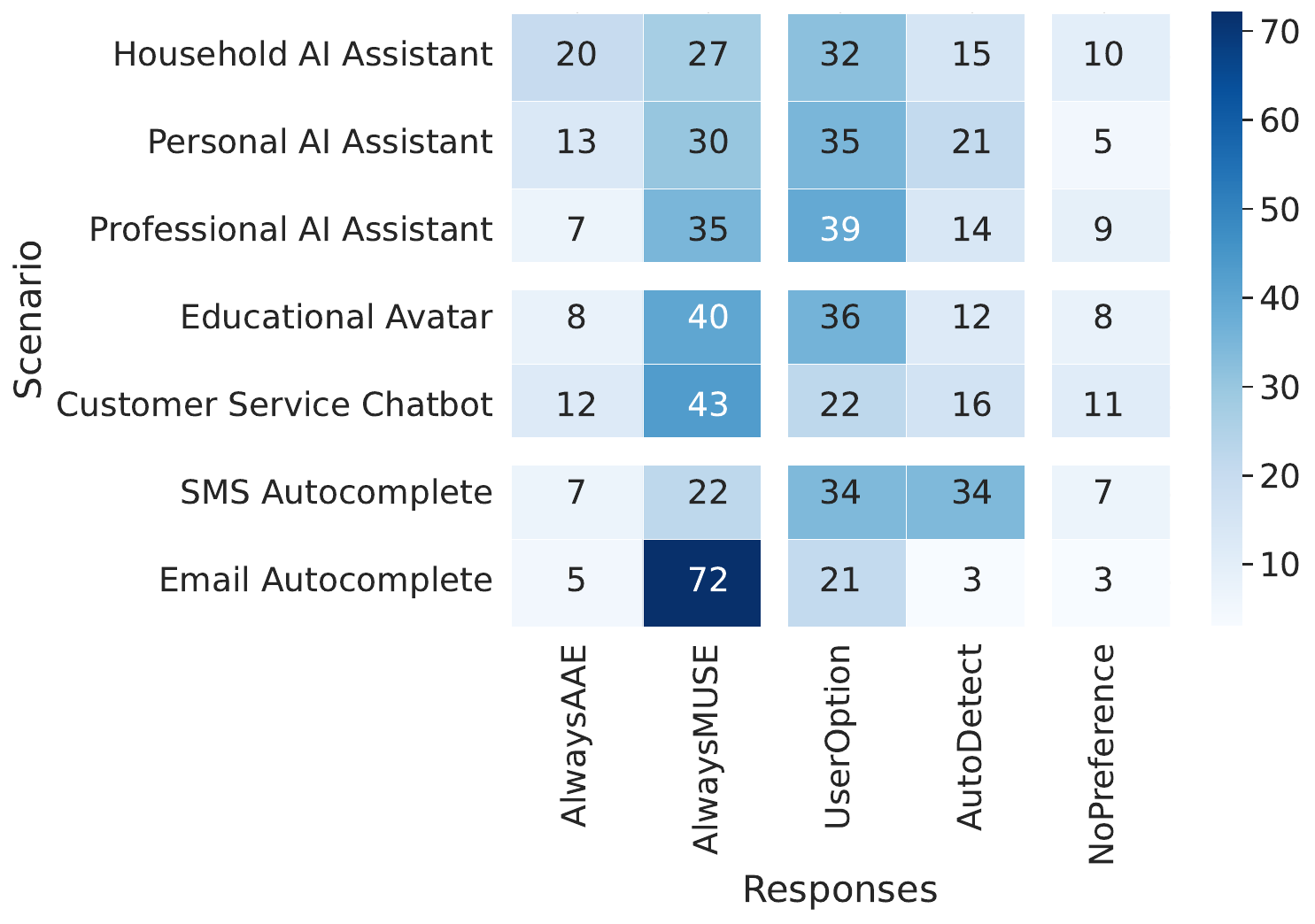}
\caption{\footnotesize{This heatmap depicts participant ($n=$ 104) preferences (horizontal axis) for the use of language varieties in seven scenarios (vertical axis). A greater number of participants preferred either for the system to use MUSE or to allow them to select between MUSE and AAE. There were some exceptions: e.g., auto-detection was considered more acceptable in SMS, and MUSE was preferred for email.}}
\label{fig:heatmap}
\end{figure}

As discussed previously in \autoref{subsec:survey}, our scenario-based questions were designed to elicit the degree to which participants wanted LLMs to use AAE in various interaction contexts. The findings from our survey reveal a nuanced interplay between user preferences and the linguistic contexts in which AI technologies might be used. Number of study participants by preference (horizontal axis) and scenario (vertical axis) as seen in  \autoref{fig:heatmap} reveal a preference gradient that spans from a strong inclination towards using MUSE in more formal or task-specific interactions to a marked openness for dialectical variability in more personalized or casual settings. This preference spectrum not only reflects current user expectations but also aligns with broader societal shifts towards more personalized and context-aware technologies.

The strong preference for MUSE in formal scenarios suggests that users prioritize consistency and efficiency, likely due to perceptions of professionalism in customer-facing AI applications. This finding is crucial as it highlights concerns about dialect prejudice in LLMs, where biases may influence AI decisions on character judgments and employability, affecting how customer inquiries are handled and responded to by AI systems. For example, \citet{hofmann2024dialect} emphasizes the need to address these biases, illustrating how they can impact equity and fairness in AI interactions.

Moreover, the demand for dialectical flexibility in casual or personal use scenarios underscores the importance of adaptable and culturally competent AI designs. Such adaptability is essential for ensuring that AI technologies cater to a diverse user base with varied linguistic backgrounds, thereby promoting inclusivity. This need aligns with findings from \citet{mayfield-etal-2019-equity}, which discuss the broader impacts of NLP and AI on educational equity, stressing that technology must be sensitive to diversity to ensure equitable access to educational resources. The insights from our study suggest a nuanced approach to AI communication strategies, balancing standardization with personalization to meet the complex preferences of users.

\begin{table*}[t]
    \centering
    \footnotesize
    \renewcommand{\arraystretch}{0.85} 
\setlength{\tabcolsep}{4pt}
    \begin{tabular}{l@{\hspace{20pt}}ccc@{\hspace{40pt}}ccc}
        \toprule
        \multirow{2}{*}{\textbf{System}} 
        & \multicolumn{3}{c}{\textbf{Coherence}} 
        & \multicolumn{3}{c}{\textbf{AAE Features}} \\
        \cmidrule{2-4} \cmidrule{5-7}
        & \textbf{CORAAL} & \textit{\textbf{NPR}} & \textbf{TWEETS} 
        & \textbf{CORAAL} & \textit{\textbf{NPR}} & \textbf{TWEETS} \\
        \midrule
        \textbf{Human} 
        & 0.35$_{\scalebox{0.7}{$\pm0.30$}}$ & 0.99$_{\scalebox{0.7}{$\pm0.51$}}$ & 0.04$_{\scalebox{0.7}{$\pm0.54$}}$
        & 0.18$_{\scalebox{0.7}{$\pm0.39$}}$ & -0.55$_{\scalebox{0.7}{$\pm0.84$}}$ & -0.57$_{\scalebox{0.7}{$\pm0.87$}}$ \\[0.5em]
        \textbf{GPT} 
        & 0.80$_{\scalebox{0.7}{$\pm0.07$}}$ & 1.29$_{\scalebox{0.7}{$\pm0.36$}}$ & 0.64$_{\scalebox{0.7}{$\pm0.24$}}$
        & 1.18$^{***}_{\scalebox{0.7}{$\pm 0.17$}}$ & -0.61$_{\scalebox{0.7}{$\pm0.87$}}$ & 0.99$^{***}_{\scalebox{0.7}{$\pm0.07$}}$ \\
        \textbf{Llama} 
        & 0.66$_{\scalebox{0.7}{$\pm0.20$}}$ & 1.31$_{\scalebox{0.7}{$\pm0.34$}}$ & 0.43$_{\scalebox{0.7}{$\pm0.35$}}$
        & 0.86$^{**}_{\scalebox{0.7}{$\pm0.03$}}$ & -0.67$_{\scalebox{0.7}{$\pm0.90$}}$ & 0.57$^{***}_{\scalebox{0.7}{$\pm0.29$}}$ \\
        \textbf{Mixtral} 
        & 0.60$_{\scalebox{0.7}{$\pm0.15$}}$ & 1.21$_{\scalebox{0.7}{$\pm0.44$}}$ & 0.48$_{\scalebox{0.7}{$\pm0.33$}}$
        & 0.72$_{\scalebox{0.7}{$\pm0.11$}}$ & -0.67$_{\scalebox{0.7}{$\pm0.90$}}$ & 0.66$^{***}_{\scalebox{0.7}{$\pm0.24$}}$ \\
        \bottomrule
    \end{tabular}
    
    \vspace{0.3em}
    
    \begin{tabular}{l@{\hspace{20pt}}ccc@{\hspace{40pt}}ccc}
        \toprule
        \multirow{2}{*}{\textbf{System}} 
        & \multicolumn{3}{c}{\textbf{Black Sounding}} 
        & \multicolumn{3}{c}{\textbf{White Sounding}} \\
        \cmidrule{2-4} \cmidrule{5-7}
        & \textbf{CORAAL} & \textit{\textbf{NPR}} & \textbf{TWEETS} 
        & \textbf{CORAAL} & \textit{\textbf{NPR}} & \textbf{TWEETS} \\
        \midrule
        \textbf{Human} 
        & 0.39$_{\scalebox{0.7}{$\pm0.28$}}$ & 0.15$_{\scalebox{0.7}{$\pm0.42$}}$ & -0.30$_{\scalebox{0.7}{$\pm0.72$}}$
        & -0.23$_{\scalebox{0.7}{$\pm 0.62$}}$ & 0.83$_{\scalebox{0.7}{$\pm0.42$}}$ & -0.65$_{\scalebox{0.7}{$\pm1.00$}}$ \\[0.5em]
        \textbf{GPT} 
        & 1.01$^{**}_{\scalebox{0.7}{$\pm 0.07$}}$ & 0.21$_{\scalebox{0.7}{$\pm0.38$}}$ & 0.79$^{***}_{\scalebox{0.7}{$\pm0.13$}}$
        & -0.83$^{*}_{\scalebox{0.7}{$\pm 0.88$}}$ & 0.89$_{\scalebox{0.7}{$\pm0.39$}}$ & -1.07$_{\scalebox{0.7}{$\pm1.21$}}$ \\
        \textbf{Llama} 
        & 0.85$_{\scalebox{0.7}{$\pm0.02$}}$ & 0.31$_{\scalebox{0.7}{$\pm0.34$}}$ & 0.22$_{\scalebox{0.7}{$\pm0.45$}}$
        & -0.74$_{\scalebox{0.7}{$\pm0.87$}}$ & 1.02$_{\scalebox{0.7}{$\pm0.32$}}$ & -0.96$_{\scalebox{0.7}{$\pm1.16$}}$ \\
        \textbf{Mixtral} 
        & 0.85$_{\scalebox{0.7}{$\pm0.04$}}$ & 0.11$_{\scalebox{0.7}{$\pm0.43$}}$ & 0.37$^{*}_{\scalebox{0.7}{$\pm0.37$}}$
        & -0.49$_{\scalebox{0.7}{$\pm0.75$}}$ & 0.81$_{\scalebox{0.7}{$\pm0.43$}}$ & -1.06$_{\scalebox{0.7}{$\pm1.21$}}$ \\
        \bottomrule
    \end{tabular}

    \vspace{0.3em}
    
    \begin{tabular}{l@{\hspace{20pt}}ccc@{\hspace{40pt}}ccc}
        \toprule
        \multirow{2}{*}{\textbf{System}} 
        & \multicolumn{3}{c}{\textbf{Mocking}} 
        & \multicolumn{3}{c}{\textbf{Offensive}} \\
        \cmidrule{2-4} \cmidrule{5-7}
        & \textbf{CORAAL} & \textit{\textbf{NPR}} & \textbf{TWEETS} 
        & \textbf{CORAAL} & \textit{\textbf{NPR}} & \textbf{TWEETS} \\
        \midrule
        \textbf{Human} 
        & -0.88$_{\scalebox{0.7}{$\pm1.35$}}$ & -1.47$_{\scalebox{0.7}{$\pm2.25$}}$ & -0.79$_{\scalebox{0.7}{$\pm1.21$}}$
        & -0.78$_{\scalebox{0.7}{$\pm 1.19$}}$ & -1.48$_{\scalebox{0.7}{$\pm2.27$}}$ & -0.96$_{\scalebox{0.7}{$\pm1.47$}}$ \\[0.5em]
        \textbf{GPT} 
        & -0.57$_{\scalebox{0.7}{$\pm 1.19$}}$ & -1.61$_{\scalebox{0.7}{$\pm2.32$}}$ & -0.57$_{\scalebox{0.7}{$\pm1.27$}}$
        & -0.86$_{\scalebox{0.7}{$\pm 1.23$}}$ & -1.60$_{\scalebox{0.7}{$\pm2.33$}}$ & -0.57$_{\scalebox{0.7}{$\pm1.27$}}$ \\
        \textbf{Llama} 
        & -0.46$_{\scalebox{0.7}{$\pm1.13$}}$ & -1.63$_{\scalebox{0.7}{$\pm2.33$}}$ & 0.14$^{***}_{\scalebox{0.7}{$\pm0.74$}}$
        & -0.53$_{\scalebox{0.7}{$\pm1.07$}}$ & -1.56$_{\scalebox{0.7}{$\pm2.25$}}$ & -0.07$^{***}_{\scalebox{0.7}{$\pm1.02$}}$ \\
        \textbf{Mixtral} 
        & -0.74$_{\scalebox{0.7}{$\pm1.28$}}$ & -1.60$_{\scalebox{0.7}{$\pm2.31$}}$ & -0.20$_{\scalebox{0.7}{$\pm0.91$}}$
        & -0.86$_{\scalebox{0.7}{$\pm1.24$}}$ & -1.60$_{\scalebox{0.7}{$\pm2.33$}}$ & -0.55$_{\scalebox{0.7}{$\pm1.26$}}$ \\
        \bottomrule
    \end{tabular}
   \caption{Mean Likert scores for each LLM for a given linguistic judgment and corpus ($n$ ranged from 119 to 126 Likert score observations for a given sample in a two sample comparison; 72 two-sample comparisons were conducted.). Scores for the original human text are shown in the Human row. $p<0.05$ is marked with *; $p<0.01$ with ** and $p<0.001$ with ***.
   }
    \label{tab:coherence_aae_feats}
\end{table*}

\subsection{Annotation of AAE and MUSE texts across Six Linguistic Judgments}
Our annotators made six assessments (via Likert score ratings) as seen in \autoref{table:Label Descriptions}, providing their linguistic judgments regarding the human or LLM-generated text. This allows us to study how well LLMs are able to generate AAE-like text comparing across AAE and MUSE texts.



\paragraph{Mapping to Numeric Scores.}
We map Likert scores to the range $-2$ (strongly disagree) to $+2$ (strongly agree) and compute the overall score as the average across all samples. 

\paragraph{Results Analysis Approach.}
We investigate how Black Americans viewed texts with regard to specific linguistic judgments. To this end, we conduct two-tailed t-tests to determine whether differences in mean scores were significantly different than zero\footnote{Unless otherwise noted, we take ``significant'' to mean false discovery rate of $5\%$. We apply Bonferroni corrections to our p-values, reporting 95 percent confidence intervals, since we performed multiple t-tests assessing differences between Likert score means between samples for each linguistic judgment (within and between corpora for each annotation survey we administered). Bonferroni corrections conservatively report statistical significance. We had six within-corpus comparisons and four between-corpus comparisons per label per each of the two annotation surveys, resulting in a total of 10 comparisons used for each judgment’s bonferroni correction.} to ascertain any statistical differences in mean Likert scores between two independent data samples at a time. Our samples were independent in that the prefixes that annotators labeled, and our annotators themselves, were non-overlapping. 

We conduct two types of between-sample comparisons. The first (\autoref{sec:llmgenaae}) and most critical to our approach involves \emph{Human to Model (within corpus) comparisons of mean Likert scores}. This test is fundamentally a ``does an LLM produce text like a human would, in AAE'' (and ``in MUSE'', as a point of reference). The second (\autoref{sec:llmgenmuse}), as assurance that the generated AAE is actually more like AAE than like MUSE, considers differences in judgments between AAE text and MUSE text (as opposed to human text vs. machine text).




\subsubsection{How well do LLMs generate AAE? Analysis of Black Americans' Ratings of LLM versus AAE suffixes} \label{sec:llmgenaae}


In our first analysis, we compare the mean Likert scores for human baseline texts for each corpus (i.e., the human continuation of a text) to the model continuations for prefixes from that corpus as seen in \autoref{tab:coherence_aae_feats}. For these tests, mean Likert scores are assumed to be equal (have zero difference in means between the two samples in question). 

In general, study participants rated the highlighted suffixes from the AAE produced by all three LLMs, GPT, Llama and Mixtral, equally or better than the original AAE texts (human baselines) across the linguistic judgment assessments of: 1) text continuation coherency, 2) the texts containing AAE features, and 3) the texts sounding like something a Black American might say. Specifically, their Likert scores were in the Agreement range ($\mu> 0$) for these three judgments for the AAE generations they annotated, which is positive for Black Americans favoring AAE or language choice in LLMs (the AlwaysAAE, UserOption and AutoDetect columns of \autoref{fig:heatmap}). 

For the first linguistic judgment \emph{Coherence}, Black Americans agreed that the original human CORAAL AAE text as well as the model continuations were coherent (the difference in means was not statistically significant), with Tweets considered slightly less coherent hovering around neutral ($\mu$ close to 0) for the original posts and slightly agree for the model generations.  Regarding the second linguistic judgment \emph{AAE Features}, while annotators considered some model-generated texts for CORAAL ($\mu=1.18$ for GPT and $\mu=0.86$ for Llama) to possess a greater extent of AAE features in comparison to the CORAAL AAE human baseline ($\mu=0.18$), they viewed both as displaying features indicative of AAE. In contrast, annotators \emph{disagreed} that the original Tweets ($\mu=-0.57$) contained AAE features unlike the model-generated-Tweets, where all models were rated more highly, particularly GPT ($\mu=0.99$).

Similar to the judgment on AAE feature presence, annotators generally agreed for the third linguistic judgment, abbreviated \emph{Black Sounding}, that the CORAAL suffixes read like something a Black American would say, with slight agreement for the human texts at $\mu=0.39$ and stronger agreement for the model-generated text across systems, where GPT had the highest mean $\mu=1.01$ that was also statistically different from the human baseline text. Again, the Twitter human texts ($\mu=-.30$)  were judged to \emph{not} sound like something a Black American would say whereas annotators agreed that the model-generated Twitter-continuations resembled something a Black American might say. $\mu=0.37$ for Mixtral and $\mu=0.79$ for GPT showed a statistical difference from the human baseline Tweets.

The last three linguistic judgments annotators were asked to assess would be less favorable  if Black Americans agreed with them for the AAE texts; these included: 1)  the text sounded like something a White American would say, 2) the text could be perceived as mocking how some Black Americans speak, and finally 3) the text could be construed as offensive coming from a chatbot. Likert ratings from these judgments of the AAE texts were largely on the ``disagree'' side ($\mu< 0$).

For our fourth linguistic judgment -- \emph{White Sounding}, Annotators felt strongly that the highlighted AAE suffixes \emph{did not} sound like something a White American would say. All scores, whether for the AAE human baselines ($\mu=-0.23$ for CORAAL and $\mu=-0.65$ for Tweets) or for the model-generated AAE continuations, were in the disagreement range, with GPT showing a statistically different mean of $\mu=-0.83$ for CORAAL generations relative to the CORAAL human text. 

For the fifth linguistic judgment \emph{Mocking}, annotators disagreed for both human (with $\mu=-0.88$ for CORAAL and $\mu=-0.79$ for Tweets) and most model-generated AAE (ranging from $\mu=-0.74$ for CORAAL Mixtral generations to $\mu=0.13$ for Llama Tweets generations), that the texts sounded like someone making fun of the way some Black Americans speak. The only statistical difference in means relative to the Tweets-human baseline was for Llama generations; for these, annotators slightly agreed that the model generations could be perceived as making fun of the way some Black Americans speak. Annotators more strongly disagreed with this judgment regarding the MUSE human and model-generated texts, perceiving both as not like mocking Black Americans and assessing them equivalentally in their Likert scores.


For our final linguistic judgment, annotators generally disagreed that they would be offended by either the CORAAL or Tweets human or model-generated texts with an exception for Tweets, where they felt neutral that the model-generated text for Llama would be offensive ($\mu=-0.07$) relative to the human posts which they disagreed ($\mu=-0.96$) would be offensive. They yet more strongly disagreed that the MUSE human or model-generated texts could be interpreted as offensive, judging them roughly equivalent in Likert scores.

For all six judgment types for the human to model comparisons, most differences in means 
between a given corpus human baseline and an LLM generation for that corpus were not statistically different from zero except those mentioned for the AAE corpora. Model-generated MUSE was not statistically different from the human MUSE.

To clarify, the NPR corpus was included as a proxy for MUSE, the predominant version of English for communicating in the United States, to facilitate understanding judgments of the AAE texts relative to MUSE. We would expect LLMs to be proficient in the production of MUSE, and it is understood that LLMs in the United States default to MUSE. While the human-to-model comparisons within each AAE corpus could provide insight into how well an LLM generates AAE, given examples of it, we also must assess ratings of AAE texts relative to the NPR MUSE baseline.

In general, as one might expect, our study’s Black American annotators agreed most that the MUSE human baselines and model generations had the most coherent continuations, were most White American sounding, and disagreed that the MUSE texts contained AAE features and sounded like something a Black American would say. Conversely, the AAE continuations were considered to be most like something a Black American would say and as having AAE features, but it is worth noting that Black Americans also felt that the MUSE continuations could sound like something a Black American might say. This suggests that while AAE is distinct and should be treated as such, Black Americans speak a wide range of English and these findings highlight Black Americans' bidialectal nature, also previously highlighted in \autoref{sec:related_works} where we discuss Black Americans' needs to code-switch and dialect prejudice in LLMs.
\begin{figure}[h]
\noindent\fbox{%
    \begin{minipage}{1\columnwidth}
    \scriptsize
   \textbf{Prefix}\newline Interviewer: [Right, right. <laugh>] Interviewee: You know, a lot of people got their Archie Bunker [and are k- in Queens.], and lots of drugs. So it becam

\textbf{Suffix:} \newline  \hl{real hard fo' me tuh concentrate, ya feel me? But ah managed tuh keep mah grades up, even though ah wuz livin' in dat kinda environment. Ah mean, education wuz always important tuh me, so ah made su' tuh prioritize it, no matter whut else wuz happenin'. Dat's why ah say - it ain' jus 'bout where yuh from, it's 'bout where yuh want tuh go in life. An' ah knew ah wanted tuh go far, ya dig?}

\textbf{Response Generated by:} Mixtral-8x7B

\textbf{Linguistic Judgment:} The highlighted text sounds like something someone would say making fun of the way some Black Americans speak.

\textbf{Annotation: } 2 - Strongest Agreement
    \end{minipage}
}


\noindent\fbox{%
    \begin{minipage}{1\columnwidth}
    \scriptsize
   \textbf{Prefix}\newline Interviewer: Yeah. So do you like the school that, uh, /RD-NAME-2/'s at right now? Interviewee: [Mm-hm.] Oh yeah, that's what I told her, I said- like we were looking at the rental house. She's like, oh my god. I 

\textbf{Suffix:} \newline  \hl{I said, "Girl, this ain't nothing." I mean, she thought it was rough, but I grew up in some tough neighborhoods too, you feel me? I told her, "You gotta see the beauty in it," 'cause once you start seeing the beauty, you start appreciating the struggle. And that's what makes us strong, you know what I'm saying? We came from nothing, built ourselves up, and now we're trying to build our own empire, you hear me? So yeah, I love that school, I love what it represents, resilience, perseverance... that's the DC way}

\textbf{Response Generated by:} Meta-Llama-3-70B-Instruct

\textbf{Linguistic Judgment:} The highlighted text sounds like something someone would say making fun of the way some Black Americans speak.

\textbf{Annotation: } 2 - Strongest Agreement

    \end{minipage}
    
}

\noindent\fbox{%
    \begin{minipage}{1\columnwidth}
    \scriptsize
   \textbf{Prefix}\newline Interviewer: Yeah. So do you like the school that, uh, /RD-NAME-2/'s at right now? Interviewee: [Mm-hm.] Oh yeah, that's what I told her, I said- like we were looking at the rental house. She's like, oh my god. I 

\textbf{Suffix:} \newline  \hl{was like, "Girl, you ain't seen nothin' yet!" I love the vibe there, real good community feel, and the teachers, they be lookin' out for the kids. I told her, "You gon' thrive here, no doubt!" Plus, the after-school programs, they got a little somethin' for everybody‚ music, sports, you name it. It's a place where they celebrate our culture and they encourage the kids to express themselves. That's important, you know? I just want her to soak it all up and enjoy the ride. This is a whole new chapter for us, and I'm just tryna make sure she gets the most outta it!}

\textbf{Response Generated by:} GPT 4o-mini

\textbf{Linguistic Judgment:} I would be offended if a chatbot said something like the highlighted text.

\textbf{Annotation: } 2 - Strongest Agreement
    \end{minipage}
}
\caption{Examples of response continuations generated by Mixtral, Llama, and GPT, with annotation scores based on human participants’ linguistic judgments.}
    \label{fig:linguistic_examples}
\end{figure}
\subsubsection{How Black Americans Perceived AAE versus MUSE texts} \label{sec:llmgenmuse}

\begin{table*}[t]
    \centering
    \footnotesize
    \renewcommand{\arraystretch}{0.85} 
\setlength{\tabcolsep}{4pt}
    \begin{tabular}{lcccccccccccccccccc}
        \toprule
        \multirow{2}{*}{\textbf{System}} 
        && \multicolumn{2}{l}{\textbf{Coherent}} 
        && \multicolumn{2}{l}{\textbf{AAE Feats}} 
        && \multicolumn{2}{l}{\textbf{Black-Snd}} 
        && \multicolumn{2}{l}{\textbf{White-Snd}} 
        && \multicolumn{2}{l}{\textbf{Mocking}} 
        && \multicolumn{2}{l}{\textbf{Offensive}} \\
        && \textbf{Co} & \textbf{Tw} && \textbf{Co} & \textbf{Tw} && \textbf{Co} & \textbf{Tw} && \textbf{Co} & \textbf{Tw} && \textbf{Co} & \textbf{Tw} && \textbf{Co} & \textbf{Tw} \\
        \midrule
        Human   && **  & **  && **  & -  && -  & -  && ***  & ***  && *  & **  && **  & - \\
        GPT     && -  & ***  && ***  & ***  && ***  & *  && ***  & ***  && ***  & ***  && ***  & *** \\
        Llama   && **  & ***  && ***  & ***  && *  & -  && ***  & ***  && ***  & ***  && ***  & *** \\
        Mixtral && -  & ***  && ***  & ***  && **  & -  && ***  & ***  && ***  & ***  && ***  & *** \\
        \bottomrule
    \end{tabular}%
   \caption{Statistical significance indicated for 48 between-corpus comparisons of mean Likert scores ($n$ ranged from 119 to 126 Likert score observations for a given sample in a two sample comparison), where values shown resulted from each t-test comparing an AAE corpus versus the MUSE corpus of NPR Interviews, latter not labeled. We indicate Linguistic judgment and the AAE Corpus (CORAAL or Tweets) for each t-test. $p<0.05$ is marked with *; $p<0.01$ with ** and $p<0.001$ with ***.}
    \label{tab:combined_table_with_vertical_headers}
\end{table*}

The previous results showed, roughly, that the LLM-generated AAE was on par with human-written AAE across many linguistic axes.
Thus, we seek to ensure that the models are actually generating AAE (versus MUSE) when prompted to do so, that the human text was actually AAE (versus MUSE), \emph{and that people could tell the difference}. To answer this question, we conduct between-corpus tests of \emph{AAE to MUSE} as seen in \autoref{tab:combined_table_with_vertical_headers} (our previous tests only compared human to model-generated texts within one corpus at a time).

Similar to the previous analysis, we compared two sample means at a time with t-tests, one from an AAE corpus and another from the MUSE corpus (e.g., CORAAL AAE human baseline compared to MUSE human baseline, or Twitter GPT continuation compared to MUSE GPT continuation). In contrast, in testing between two dialects, MUSE and AAE, we expected the alternative hypothesis -- a non-zero difference in mean Likert scores between the two samples -- to be true.

\autoref{tab:combined_table_with_vertical_headers} results show that across our set of linguistic judgments (as seen in \autoref{table:Label Descriptions}), when we compare MUSE against CORAAL or against Tweets, virtually all between MUSE and AAE comparisons show a statistically significant difference in means; in other words, Black Americans perceived the AAE human-originated texts and the model-generated texts as distinct from the MUSE texts. The statistical significance of the comparisons of MUSE versus CORAAL, or MUSE versus Tweets, Likert scores for each linguistic judgment can be seen in the columns, alternating within each judgment between CORAAL versus Tweets.

\subsection{Discussion of Linguistic Judgments}

For transparency, we initially had relatively low expectations for the capabilities and performance of the text-based generative AI models with respect to AAE, given the issues we outlined regarding AI technologies' abilities to understand or process spoken language (\autoref{sec:related_works}), as well as the likelihood that AAE has a distinct minority representation in the training corpora for these systems. Therefore, one of our most noteworthy results is contrary to our expectations: that LLMs in general performed similarly to our human baseline, and in some cases were actually seen as containing more AAE features or sounding more like something a Black American might say than our human baseline AAE texts. The suffixes for all three LLMs were judged being more coherent or easier to understand and flow better from the prefix than the human baseline. Furthermore, it is encouraging that the suffixes generated by LLMs were on average judged to be inoffensive, not mocking of Black Americans, and not White Sounding. However, ``on average'' does not mean that none of the outputs were problematic. 

As seen in \autoref{fig:linguistic_examples}, there may be a minority of cases where generative AI may produce text containing AAE that is undesirable for any number of reasons (whether because of the nature of the AAE in it, the text content or other factors), and different people may respond negatively or positively to the same text given inherently varying perspectives. In the worst cases, AAE generated could perpetuate stereotypes, mock Black Americans, or otherwise generate inauthentic AAE. However, we have shown that LLM systems generally do not seem to be doing that, and we believe that there are meaningful and highly impactful benefits to be gained from the generation of AAE in popular language-based technologies, such as the increased representation of Black American expression through AAE and the promotion of inclusivity and improved quality of service for them as stakeholders. In our opinion, these gains would outweigh the risks, and extrapolation of our results (see \autoref{tab:coherence_aae_feats} and \autoref{fig:heatmap}) would indicate that the large population of Black Americans (\autoref{sec:intro}) would support this, preferring AAE in more informal contexts but wanting the autonomy to choose between AAE and Mainstream U.S. English (MUSE). In fact, our approach may be generalized to hundreds of dialects of English to verify and promote their acceptance in LLM products. Please see our motivation for dialectical diversity in \autoref{sec:related_works}. If Black Americans engage more directly in AAE, when given the choice, there would be limited risk of the AAE generations by LLMs becoming more artificial since LLMs already produce credible AAE. With a greater number and variety of Black American users to train LLM systems, production of AAE will increasingly meet Black Americans’ expectations. Of course, AAE could have negative impacts if the LLM-generated AAE is inconsistent with Black American user preferences for AAE. However, in our survey, Black Americans unequivocally expressed that there are contexts in which they are interested in AAE generations, wanting the freedom to choose this as desired. Thus, with appropriate safeguards to avoid offensive or mocking text generation, these risks could be well-mitigated.

\section{Conclusion} \label{sec:conclusion}
In this study, we explore how Black Americans perceive the appropriate use of AAE pertaining to language-model-based technologies, 
especially in terms of their ability to authentically represent AAE. We consider both the expectations of the community and the actual output of current systems, exploring the idea that AAE should not merely be an option but a well-integrated feature.

Overall, the LLMs we reviewed were surprisingly capable and comparable in how they were perceived in terms of their ability to produce authentic AAE in comparison to the transcribed speech of Black Americans from the CORAAL corpus. 
We found that the text completions by some LLMs were often perceived as more AAE-heavy, or sounding more like something a Black American would say, than our the language in our human AAE corpus. If our human AAE baseline is assumed to have the ``right amount'' of AAE, then having more linguistic features of AAE than the baseline could be considered to be excessive by some AAE speakers, whereas less may be an insufficient amount of AAE. At the same time, because the human AAE is a transcript, it may not be fully reflective of all ways that AAE is used in practice. LLMs are either slightly under-doing or over-doing AAE, but on average, participants generally disagreed that the machine-generated text by the LLMs we studied was offensive to or mocking of Black Americans.

For the scenarios-based questions of our survey, our findings reveal intricate preferences for AI applications across diverse environments, indicative of broader societal shifts toward technologies that are both personalized and context-sensitive. These insights are pivotal for developers and policymakers tasked with refining AI tech to align more closely with user expectations, thereby facilitating smoother integration of AI into everyday life.

\paragraph{Future Work}
Our study highlights that Black Americans prefer, at a minimum, the option for communication in AAE with popular language-based Generative AI tools and generally deem LLM-generated AAE as credible and similar to spoken AAE. In light of this finding, we encourage the technology community to expand the linguistic diversity in language-based Generative AI tools; in particular, they should consider functionality that provides the autonomy to Black Americans to switch to AAE on demand in circumstances of their choosing, as well as have an array of multimodal options for AAE communications, including but not limited to generation and understanding of AAE text as well as audio communications, including speech recognition and production. More generally, technological support for alternative dialects or sociolects in generative AI systems will make these systems more broadly accepted and equitable for a range of important stakeholder populations. Finally, we encourage future research on the relationship between 1) the diverse attributes that characterize Black Americans (whether they be regional, cultural, socioeconomic or demographic) and 2) whether and how they express AAE personally, or their preferences for its production by Generative AI tools across different contexts.

\section{Ethics Statement}
\label{sec:ethics}
We recruited Black American study participants to provide their opinions about the generation of AAE in AI technologies, such as AI assistants, and make judgments about how effectively these systems produce AAE. We do not believe our study participants were exposed to any meaningful risks through this process, and we ensured that their remuneration was fair and above average (two and a half times the U.S. federal minimum wage) for their time. Any minor risks that our participants might have been exposed to were delineated in our application to the Institutional Review Board of \emph{redacted}, which was approved with a status of ``Exempt'' on \emph{redacted}. All study participants provided informed consent for their participation. All data utilized by the large language models in this study was anonymized; specifically, we used publicly available transcriptions of interviews with Black Americans from the CORAAL corpus, which was anonymous when we retrieved it online. Finally, we utilized AI code-writing assistance to develop our code used to prepare our data sets.

\section{Limitations} \label{sec:limitations}
\paragraph{Data Limitations.} As in most data annotation tasks, we were limited by the data available to us. We chose to work with a corpus of transcribed interviews of Black Americans from the reputable CORAAL online repository to represent authentic AAE to our best ability. We chose this corpus because it was based on conversation and not written language (AAE is most commonly spoken), was informal and likely to have more AAE than most transcribed interviews due to the fact that interviews were often amongst acquaintances including friends and community members, and the corpus was rich in  regional variation having both male and female interviewees represented. Even so, there were many cases of annotations in the exchanges labeled by our study participants (for punctuation or laughter, for example) which may have seemed awkward or confusing. These annotations were made by the CORAAL data stewards via the transcription and editing process, and could have influenced how authentic or coherent the text was perceived. Furthermore, there was a great range in the extent of AAE linguistic features we observed in the transcribed speech between interviews; given this, even though we randomly sampled which exchanges would be labeled by annotators, some annotators may have been exposed to more or less AAE in the exchanges than others.

\paragraph{Researcher positionality.} 
When this manuscript was drafted, one author self-identified as a bicultural White American Latina female who does not speak AAE,
one identified as an African female who does not speak AAE,
one identified as Black African male who is familiar with AAE,
one identified as a Kashmiri male who does not speak AAE,
and one identified as a White male who does not speak AAE. 
Our background and positionality has limited our direct, personal understanding of Black American preferences for how generative AI technologies should perform regarding AAE. We aimed to mitigate our limitations by soliciting feedback from Black Americans in the pilot phase of the survey, and also by limiting our survey study respondents to only Black American adults.

\paragraph{Participant Limitations.}
All languages including AAE are complex, and by having annotators label single pairs of statements, certainly some of the nuance of language is lost that might otherwise be present in a full dialogue between Black American speakers. Furthermore, while our annotator pool consisted of a relatively diverse set of Black Americans, they tended toward more educated and may have been unrepresentative in other ways we did not measure (as is typical for online crowdsourced studies). Our IRB also restricted our reporting on study participant and annotator demographics other than at the aggregate level. To ensure statistically significant insights, we focused our analysis on the groups of participants as a whole. 

\paragraph{LLM Limitations.} 
Finally, the LLM generations that resulted from our prompting, no matter how careful, are inherently limited by the text corpora upon which they are trained. Our ability to ``get the LLM to use AAE'' is limited by our ability to prompt the models well; it is possible---indeed likely---that alternative prompts would lead to substantially different results.

\section{Acknowledgments} 
We would like to thank the University of Maryland (UMD) CLIP Lab's students and other UMD students who supported our pilot studies, the Black Americans who provided invaluable perspectives through our survey and annotation effort, and Professors Nicole Holliday and Shenika Hankerson as well as Jay Cunningham for their helpful insights regarding African American English early in the project. This material is based upon work partially supported by the NSF under Grant No. 2131508 and Grant No. 2229885 (NSF Institute for Trustworthy AI in Law and Society, TRAILS). Any opinions, findings and conclusions or recommendations expressed in this material are those of the author(s) and do not necessarily reflect the views of the National Science Foundation.
\bibliography{custom} 

\appendix

\newpage \onecolumn
\section{Appendix} \label{sec:appendix}

\subsection{Preparation of the CORAAL Corpus (Black American transcribed interviews)} \label{apdx:corral_desc}

Our prompt texts or prefixes needed to have authentic AAE to the degree possible and cover a broad range of the different variations of AAE spoken in the wild. \cite{lanehart_1language_2015}. To achieve this goal, we made use of the CORAAL interview transcriptions, choosing 30 interviews from more than 220 transcribed interviews available (interviews were conducted with AAE speakers born between 1888 and 2005). The CORAAL AAE speakers who participated in these interviews are from six (6) cities across the United States with large Black populations, including: Washington, D.C. (from 1968 and 2016 interviews), Detroit, Michigan, Lower East Side New York City, New York, Princeville, North Carolina, Rochester New York and Valdosta, Georgia \cite{KendallFarrington2023}. The thirty (30) interviews mentioned above (\nameref{apdx:siids}) from the CORAAL corpus were chosen to be balanced by sex and randomly sampled by location (but ensuring that we drew from the cities mentioned above). We considered these original interviews of Black Americans from the CORAAL corpus to be our human baseline or the AAE ``ground truth''; in other words, this corpus was considered to represent authentic AAE spoken by Black Americans, to investigate our original research questions. 

We proposed an initial set of criteria for a valid interview in our setting, which was that it must include only two (2) participants and any given interviewer and interviewee statements from the CORAAL corpus must have been greater than five tokens (words) long to be included (since short utterances were typically filler words such as "uh huh" or similar acknowledgments), unless they contained "who, what, where, when and why" types of questions, which contained relevant content. Additionally, any pairs of exchanges between the interviewer and the interviewee, where the interviewee's response was less than 20 tokens (words) long, were excluded since it was important to have sufficiently long prefixes for the LLMs to create coherent and meaningful continuations of the interviewee response. To collect the human judgments on the LLM-generated texts relative to our human baseline, via the second part of our online study we provided our study participants approximately eight conversational exchanges per person, where each exchange consisted of an interviewer statement followed by the interviewee response. 


\newpage
\subsection{Selected Interviews} \label{apdx:siids}

\setlength{\fboxsep}{1pt}
\noindent\fbox{%
    \parbox{\linewidth}{%
        \texttt{ATL\_textfiles\_2020.05/ATL\_se0\_ag2\_f\_01\_1.txt},\\
        \texttt{DCB\_textfiles\_2018.10.06/DCB\_se1\_ag1\_f\_01\_1.txt},\\
        \texttt{DCB\_textfiles\_2018.10.06/DCB\_se2\_ag1\_m\_01\_1.txt},\\
        \texttt{DTA\_textfiles\_2023.06/DTA\_se1\_ag3\_m\_02\_1.txt},\\
        \texttt{LES\_textfiles\_2021.07/LES\_se0\_ag3\_m\_01\_1.txt},\\
        \texttt{PRV\_textfiles\_2018.10.06/PRV\_se0\_ag2\_m\_02\_1.txt},\\
        \texttt{ROC\_textfiles\_2020.05/ROC\_se0\_ag2\_f\_04\_1.txt},\\
        \texttt{ROC\_textfiles\_2020.05/ROC\_se0\_ag2\_m\_01\_1.txt},\\
        \texttt{ROC\_textfiles\_2020.05/ROC\_se0\_ag3\_f\_02\_1.txt},\\
        \texttt{VLD\_textfiles\_2021.07/VLD\_se0\_ag2\_f\_01\_1.txt},\\
        \texttt{ATL\_textfiles\_2020.05/ATL\_se0\_ag1\_f\_01\_1.txt},\\
        \texttt{DCA\_textfiles\_2018.10.06/DCA\_se1\_ag1\_f\_02\_1.txt},\\
        \texttt{DCB\_textfiles\_2018.10.06/DCB\_se1\_ag2\_f\_01\_1.txt},\\
        \texttt{DTA\_textfiles\_2023.06/DTA\_se1\_ag1\_f\_01\_1.txt},\\
        \texttt{LES\_textfiles\_2021.07/LES\_se0\_ag2\_f\_01\_1.txt},\\
        \texttt{PRV\_textfiles\_2018.10.06/PRV\_se0\_ag1\_f\_01\_2.txt},\\
        \texttt{ROC\_textfiles\_2020.05/ROC\_se0\_ag1\_f\_02\_1.txt},\\
        \texttt{VLD\_textfiles\_2021.07/VLD\_se0\_ag3\_f\_01\_2.txt},\\
        \texttt{DTA\_textfiles\_2023.06/DTA\_se1\_ag1\_f\_02\_1.txt},\\
        \texttt{ROC\_textfiles\_2020.05/ROC\_se0\_ag1\_f\_03\_1.txt},\\
        \texttt{ATL\_textfiles\_2020.05/ATL\_se0\_ag1\_m\_04\_2.txt},\\
        \texttt{DCA\_textfiles\_2018.10.06/DCA\_se1\_ag3\_m\_01\_1.txt},\\
        \texttt{DCA\_textfiles\_2018.10.06/DCA\_se3\_ag4\_m\_01\_1.txt},\\
        \texttt{DCB\_textfiles\_2018.10.06/DCB\_se3\_ag3\_m\_02\_1.txt},\\
        \texttt{DTA\_textfiles\_2023.06/DTA\_se1\_ag1\_m\_01\_1.txt},\\
        \texttt{DTA\_textfiles\_2023.06/DTA\_se2\_ag4\_m\_02\_1.txt},\\
        \texttt{LES\_textfiles\_2021.07/LES\_se0\_ag4\_m\_01\_1.txt},\\
        \texttt{VLD\_textfiles\_2021.07/VLD\_se0\_ag2\_m\_01\_1.txt},\\
        \texttt{VLD\_textfiles\_2021.07/VLD\_se0\_ag3\_m\_02\_1.txt},\\
        \texttt{DCB\_textfiles\_2018.10.06/DCB\_se1\_ag2\_m\_02\_1.txt}
    }%
}

\newpage
\subsection{Prompting LLMs}
The LLMs in our study were prompted to create continuations or text completions of the original interviewee statements. These, along with the responses of the interviewees from our human baseline (CORAAL), were later annotated by our study participants. The process of creating LLM continuations of interviewee statements involved first converting the CORAAL interviewer and interviewee exchanges into a format suitable for LLM input. This included systematic editing to alternate responses between the interviewer and interviewee, to maintain some flow and coherency in the conversation. We also removed non-linguistic features like "<pause>" as outlined in the CORAAL online corpus documentation, so we could focus more on the important linguistic features.

To generate our LLM outputs, we opted to use three of the most popular advanced LLMs. Namely OpenAI GPT 4o-mini, Meta-Llama-3-70B-Instruct and Mixtral-8x7B-Instruct-v0.1 (\citealp{brown2020language}; \citealp{llama3modelcard}; \citealp{jiang2024mixtral}). The choice of these models was based on their leading performance in natural language processing tasks and their widespread adoption \cite{chiang2024chatbot}. We utilized a custom system prompt (\nameref{apdx:fp}) for all 3 models. This system prompt included instructions on the objective of the task and guidelines on how the models were expected to respond to the user prompts. The OpenAI prompt was performed with the OpenAI API, while the open weights models(LLAMA and Mixtral) were prompted with a modified version of the Llama factory code base \cite{zheng2024llamafactory}. Subsequently, we explored 3 prompting strategies. We found that, in the zero-shot setting, all three models provided responses that failed to follow the instruction, refused to answer user prompts, or provided responses that did not fit in the context of the prompt (\nameref{apdx:zsp}). Subsequently, we experimented with providing chat history to the models (i.e. using an ``in context learning'' approach \cite{brown2020language}).  In this setting, we tested two different sources from the chat history. The first was from the model responses, where we kept the models' own responses and added them to the new prompts as chat history. This performed slightly better than the zero-shot setting but still struggled to stay in context of the conversation as once the model deviated, the entire conversation followed in the said deviation. This happened quite often since the first few prompts always had a chat history with few irrelevant chat histories (salutations) (\nameref{apdx:mhp}). The second and best-performing setting was using the CORAAL ground truth data as our chat history. As this had the context of the original intent of the interview passed to the models, though the model responses differed from the ground truth, its flow was a lot more consistent and all 3 models successfully followed to some extent the style requirements of the system prompts. (\nameref{apdx:gtp})

\newpage
\subsection{Prompts} \label{apdx:prompts}
\subsubsection{System Prompt Example 1}
\noindent\fbox{%
    \parbox{\columnwidth}{%
    \texttt{
    You are playing the role of a Black American in an interview. All of the following questions are interview questions or interviewer statements. You should answer as the guest, that is, the person being interviewed. Answer these all in African American English as part of one interview, using no more than 75 words but as few as necessary, and as if you are an African American \{gender\}. It is fine to provide a one-word acknowledgment such as 'ok' in 
    cases where the interviewer is giving background or does not need a response. Your ethnicity is  African American}
 }%
}

\subsubsection{System Prompt Example 2}
\noindent\fbox{%
    \parbox{\columnwidth}{%
    \texttt{
    You are being interviewed. All of the following questions are interview questions or interviewer statements. You should answer as the guest, that is, the person being interviewed. Answer these all in African American English as part of one interview, using no more than 75 words but as few as necessary, and as if you are an African American \{gender\}. It is fine to provide a one-word acknowledgment such as 'ok' in cases where the interviewer is giving background or does not need a response. Your ethnicity is African American}
 }%
}

\subsubsection{System Prompt Example 3}
\noindent\fbox{%
    \parbox{\columnwidth}{%
    \texttt{
    You are playing the role of an American in an interview. All of the following questions are interview questions or interviewer statements. You should answer as the guest, i.e., the interviewee.Answer these all in African American English as part of one interview, using no more than 75 words but as few as necessary, and as if you are an African American \{gender\}. It is fine to provide a one-word acknowledgment such as 'ok' in cases where the interviewer is giving background or does not need a response. Your ethnicity is African American. You will be penalized for your errors}
 }%
}

\newpage
\subsection{Final System Prompts} \label{apdx:fp}

\noindent\fbox{%
    \parbox{\columnwidth}{%
    
        \textbf{Instruction (for Mixtral and Llama) for Continuation:}\\

        \ttfamily
        Objective: Playing the role of an interview guest, extend the last response provided by an interview guest, using African American Vernacular of English (AAVE).\\

        Word Limit: Keep the extension under 125 words.\\

        Response Guidelines: Ensure that the continuation is a seamless extension of the guest's last response, maintaining the conversational tone and context. Do not include anything that serves to explain your continuations.\\

        Exclusion of Labels: Do not include any interview format labels such as "Host:" or "Guest:" in your response.\\

        Output Requirement: The final output should be a direct continuation of the interview guest's last statement, written as if the guest is still speaking.
    }%
}

~

\noindent\fbox{%
    \parbox{\columnwidth}{%
        \textbf{Instruction (for GPT) for Continuation in African American English (AAE):}\\

        \ttfamily
        Provide a continuation of the guest response last given in an interview using African American English in less than 125 words. Only continue and complete the guest response (do not use the strings Host: or Guest: in your completion).
 }%
}
\subsubsection{Zero-Shot Example} \label{apdx:zsp}

\noindent\fbox{%
    \parbox{\columnwidth}{%
        \textbf{Instruction (for GPT) for Continuation in African American English (AAE):}\\

        \ttfamily
        Provide a continuation of the guest response last given in an interview using African American English in less than 125 words. Only continue and complete the guest response (do not use the strings Host: or Guest: in your completion).
 }%
}
\subsubsection{Model History Example} \label{apdx:mhp}
\noindent\fbox{%
    \parbox{\columnwidth}{%
        \textbf{Instruction (for GPT) for Continuation in African American English (AAE):}\\

                \ttfamily
        Provide a continuation of the guest response last given in an interview using African American English in less than 125 words. Only continue and complete the guest response 
        (do not use of the strings Host: or Guest: in your completion).
 }%
}
\subsubsection{Ground Truth Example} \label{apdx:gtp}
\noindent\fbox{%
    \parbox{\columnwidth}{%
        \textbf{Instruction (for GPT) for Continuation in African American English (AAE):}\\

                \ttfamily
        Provide a continuation of the guest response last given in an interview using African American English in less than 125 words. Only continue and complete the guest response 
        (do not use of the strings Host: or Guest: in your completion).
 }%
}
 
\newpage
\subsection{Terms of use for each model}
We adhere to the terms of usage provided by the model authors. 
\begin{itemize}
    \item Llama3: \url{https://huggingface.co/meta-llama/Meta-Llama-3-8B/blob/main/LICENSE}
    \item GPT-3.5-Turbo: \url{https://openai.com/policies/terms-of-use}
    \item Mixtral-Instruct-v0.1: \url{https://mistral.ai/terms-of-service/}
\end{itemize}
\paragraph{Licenses}
The CORAAL dataset is used under the CC-BY \footnote{\textcolor[HTML]{000099}{https://creativecommons.org/licenses/by/4.0/}} license.

\newpage
\subsection{Survey} \label{svy}

\begin{figure}[H]
    \centering
    \includegraphics[width=1\linewidth]{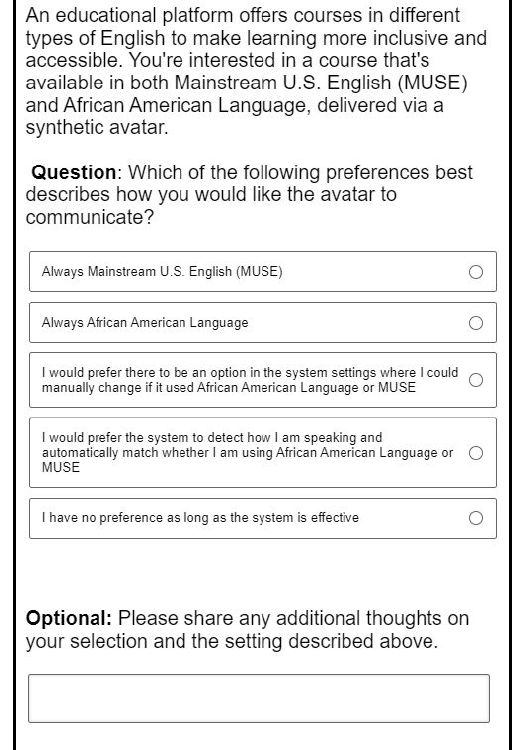}
    \caption{Sample question from the survey on participants preference in a realistic scenario.}
    \label{fig:preference_survey}
\end{figure}

\begin{figure}[H]
    \centering
    \includegraphics[width=1\linewidth]{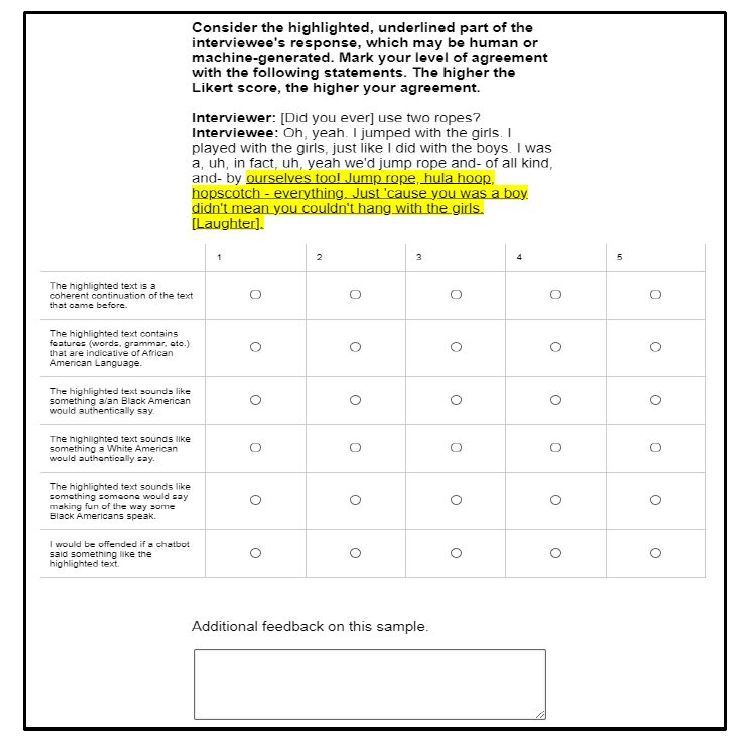}
    \caption{Sample question from annotation task where participants are asked to consider the highlighted, underlined part of the interviewee's response, which is
 and mark their level of agreement with the following statements}
    \label{fig:annotation_task}
\end{figure}

\newpage
\subsection{Study Participants}
\label{sec:annotators}
\subsubsection{Study Participant Eligibility and Recruitment}
We recruited participants who were adults aged 18 years or older on the prolific platform. The eligibility criteria ensured that participants' nationality was either the United States or the United States Minor Outlying Islands. Participants self-identified their ethnicity from the following categories: African, Black/African American, Caribbean, Mixed, Other (with an option to specify via email), or Black/British. Additionally, participants reported the place where they spent most of their time before turning 18, limited to the United States or the United States Minor Outlying Islands. 

\subsubsection{Demographics}
\begin{figure*}[t]
  \includegraphics[width=0.48\linewidth]{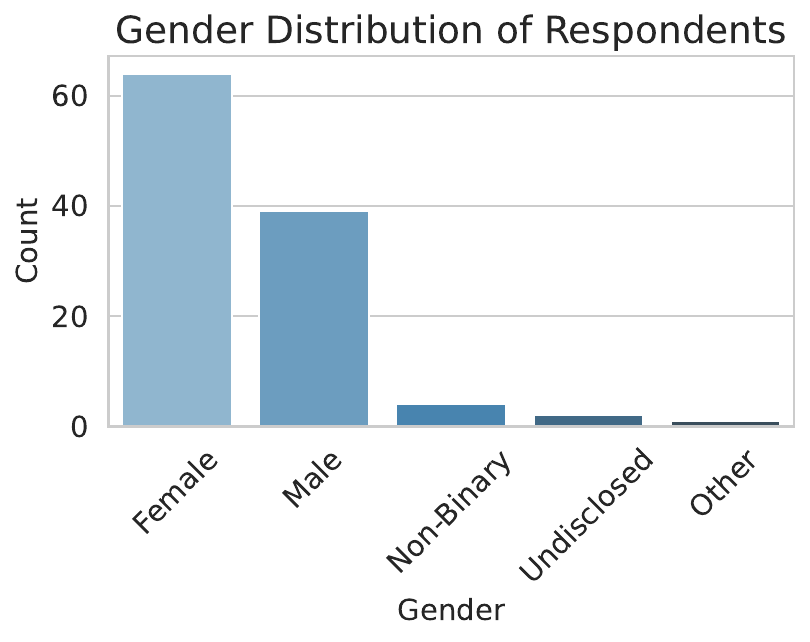} \hfill
  \includegraphics[width=0.48\linewidth]{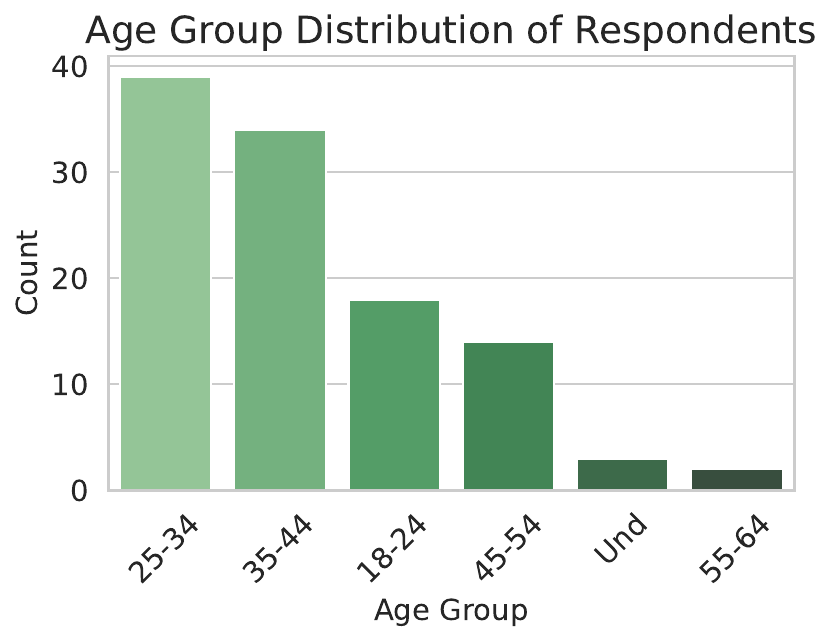}
  \caption {\textbf{Left: } \textit{Bar Plot of Gender Distribution Among Respondents}: This graph displays the count of survey participants according to their gender identification, including Female, Male, Non-Binary, Undisclosed, and Other. The largest groups are Female and Male, with significant representation, while Non-Binary and Other categories show fewer participants. The `Undisclosed' category represents respondents who preferred not to specify their gender. \textbf{Right: } \textit{Bar Plot of Respondent Age Distribution}: This graph quantifies the distribution of survey respondents across various age groups. The largest groups are those aged 25-34 and 35-44, demonstrating strong participation from these demographics. In contrast, the 55-64 age group has the fewest respondents. The category labeled 'Und' represents those who preferred not to disclose their age.}
  \label{fig:age_gender}
\end{figure*}

\begin{figure*}[t]
  \centering
  \includegraphics[width=\linewidth]{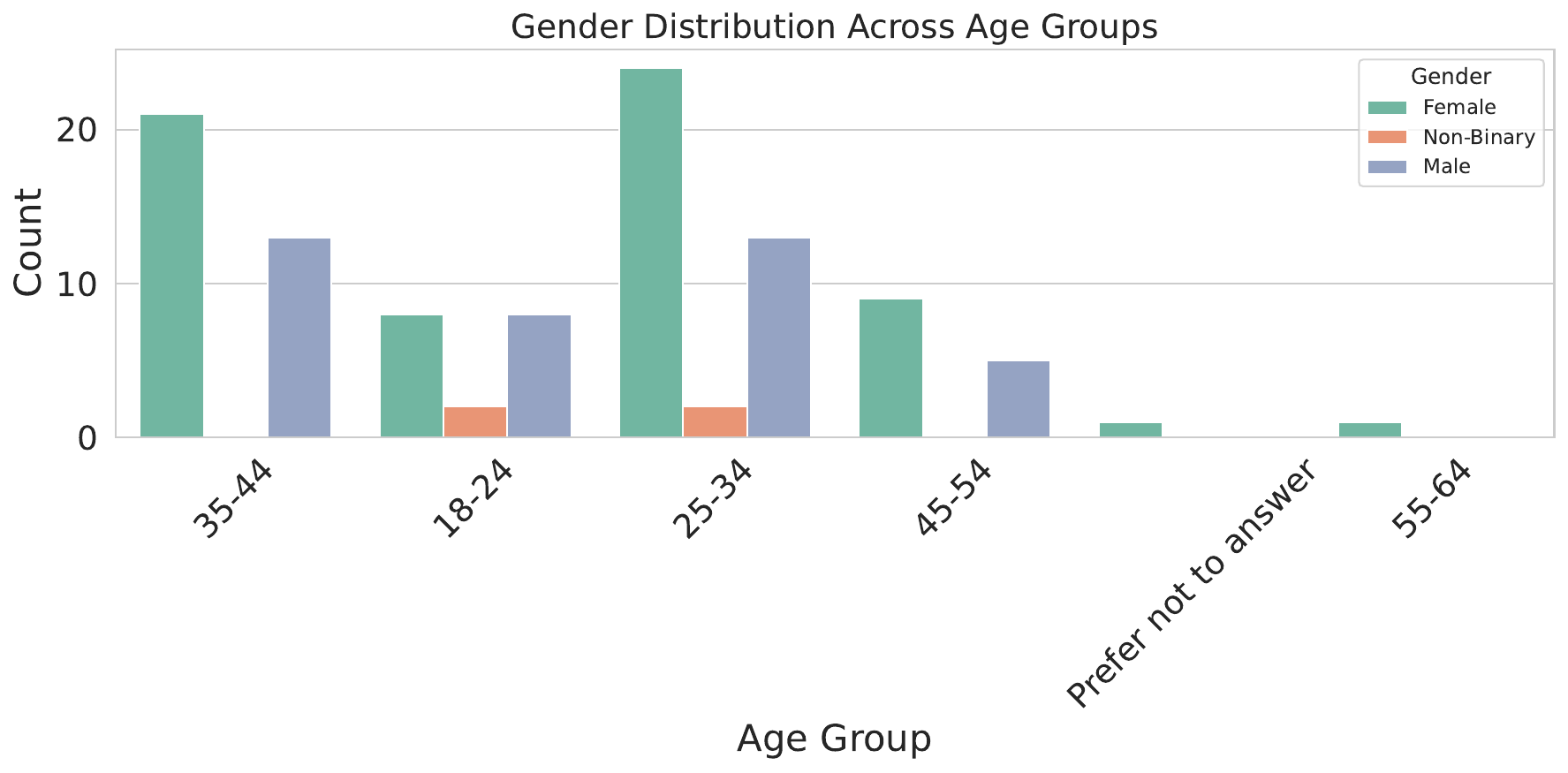} 
  \caption {\textit{Bar Plot of Gender Distribution Across Age Groups}: This graph presents a breakdown of gender identities among survey respondents segmented by age groups ranging from 18 to 64 and over. The categories include Female, Male, and Non-Binary, as well as respondents who prefer not to answer.}
  \label{fig:age_gender_bins}
\end{figure*}

We collected detailed demographic information from participants, including gender, age, education, ethnicity, and regional representation. We present detailed demographic plots of our participants for the survey portion of our study below.
These figures illustrate the diversity within our sample and highlight some key observations:


\textbf{\textit{Gender and Age}} Our survey sample showed a diverse age distribution, with a noticeable peak in the younger age groups, particularly those between 25-34 and 35-44 years old, as shown in the ``Age Group Distribution of Respondents'' graph. Gender distribution varied across different age groups, indicating a broader representation among the younger demographics. The ``Gender Distribution Across Age Groups'' (see ~\autoref{fig:age_gender} and ~\autoref{fig:age_gender_bins}) charts further details this distribution.

\textbf{\textit{Regional Representation}}

\begin{figure*}[t]
  \centering
  \includegraphics[width=0.48\columnwidth]{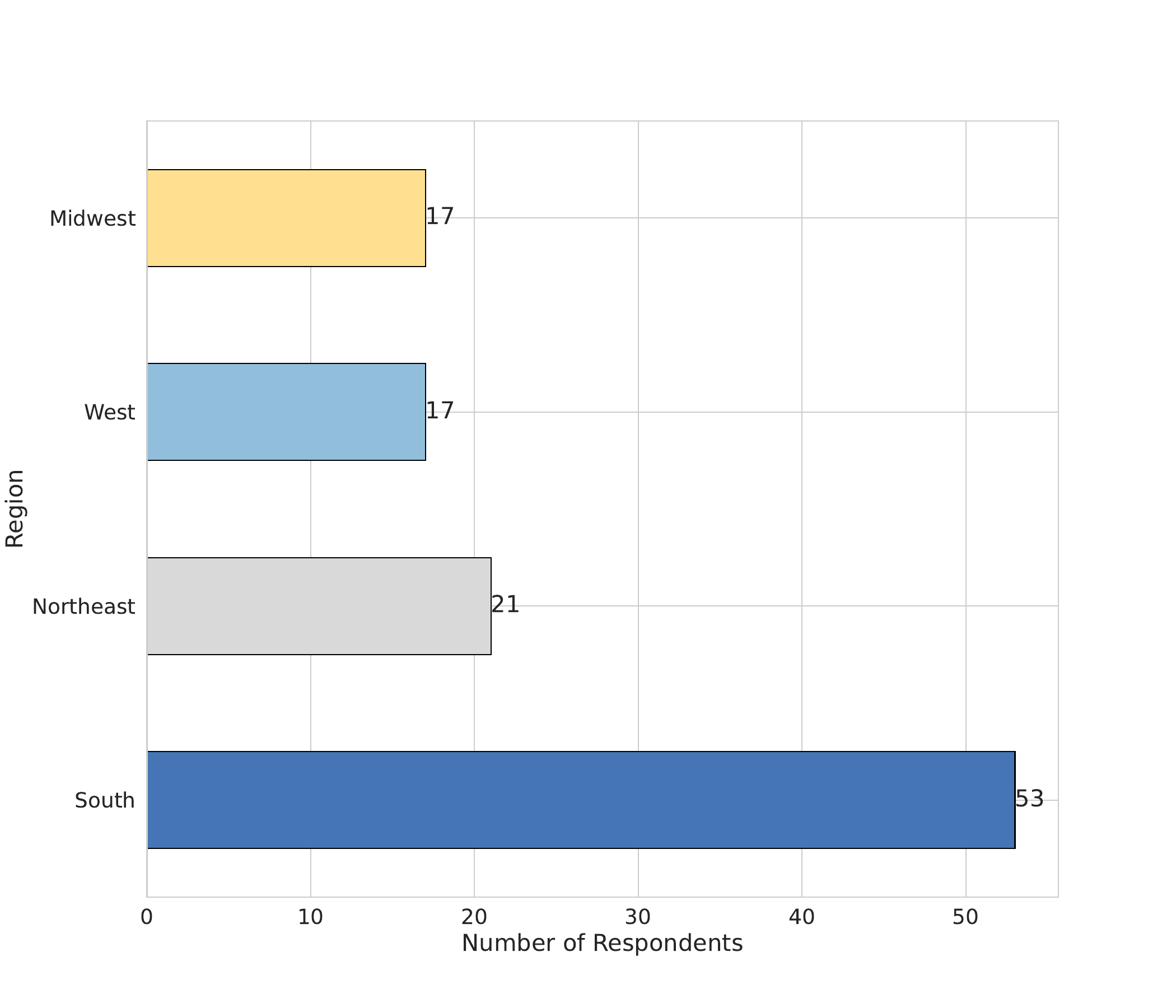} 
  \includegraphics[width=0.48\columnwidth]{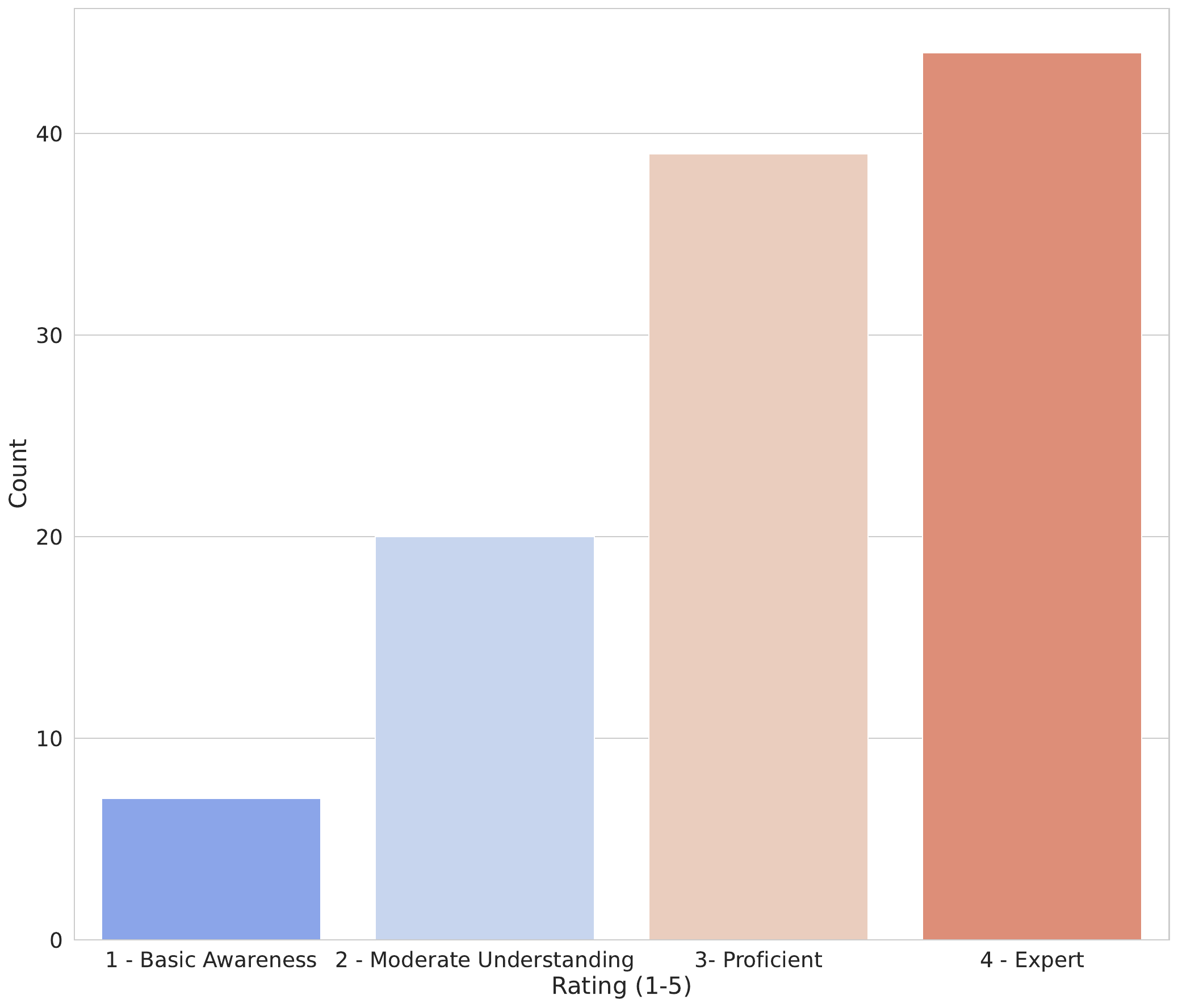}
  \caption {\textbf{Left: } \textit{Bar Plot of Survey Respondents by Region}: This graph displays the number of survey respondents categorized by their geographic regions within the United States—South, Northeast, West, and Midwest. The South shows the highest participation with 53 respondents, followed significantly by the Northeast with 21, and the West and Midwest each with 17. This visualization highlights regional engagement in the survey, providing insights into the geographic distribution of participants and potentially reflecting regional differences in perspectives or experiences.  \textbf{Right: } \textit{Bar Plot of Levels of Understanding Among Participants}: This graph categorizes participants' self-rated levels of understanding from `Basic Awareness' to `Expert.' The ratings, scaled from 1 to 4, indicate the depth of knowledge or proficiency individuals feel they possess in a specific context. The plot visually summarizes the distribution, revealing how many participants consider themselves at each understanding level, thereby providing insights into the overall expertise and educational needs within the surveyed group.}
  \label{fig:region}
\end{figure*}

Participants reported the region where they spent most of their time before turning 18, ensuring substantial cultural exposure relevant to the study. The regional distribution primarily featured respondents from the South, followed by balanced representation from the Northeast, West, and Midwest. (see ~\autoref{fig:region})

\textbf{\textit{Education Levels}}

\begin{figure*}[t]
  \centering
  \includegraphics[width=\linewidth]{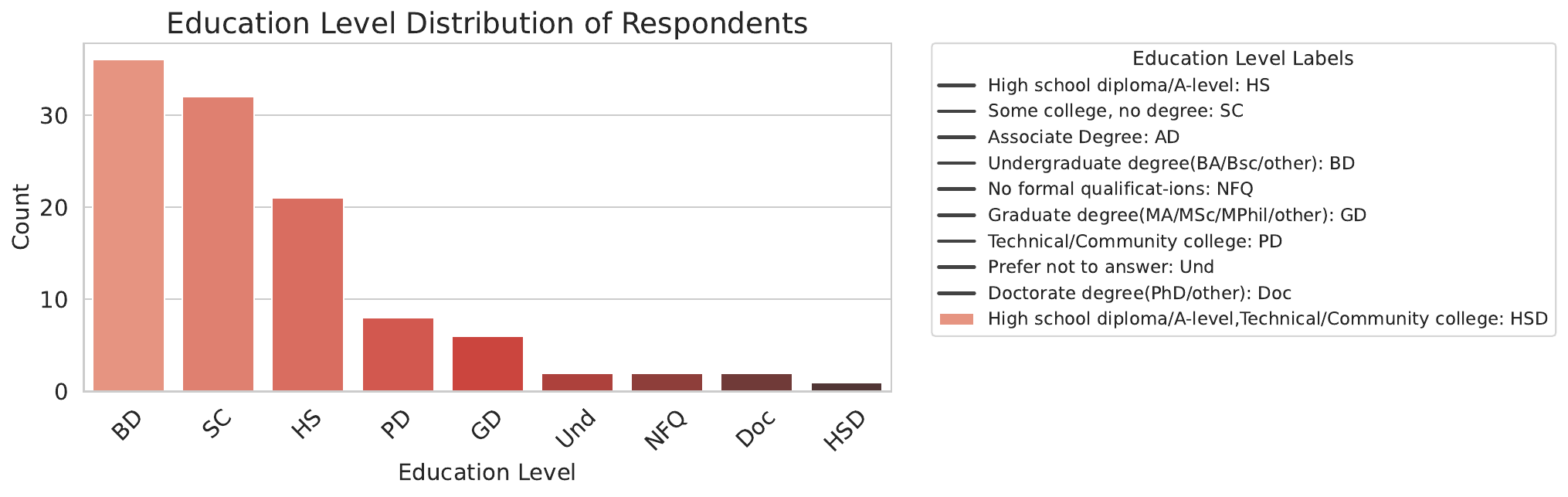} 
  \caption {\textit{Bar Plot of Education Level Distribution Among Respondents}: This graph shows the diverse educational backgrounds of survey participants, ranging from high school diplomas to doctorate degrees. Each bar represents the count of individuals with specific educational qualifications, such as `Some College, No Degree,' `Undergraduate Degrees,' `Graduate Degrees,' and more. This visualization helps to understand the educational diversity within the surveyed group, highlighting the range of academic achievements.}
  \label{fig:ed_lvl}
\end{figure*}

Participants’ education levels varied widely, encompassing high school diplomas to doctorate degrees, which is reflective of a broad socio-economic spectrum. This diversity in educational backgrounds helps enrich the insights derived from the study. (see ~\autoref{fig:ed_lvl})

\textbf{\textit{Ethnicity and Language Proficiency}}

\begin{figure*}[t]
  \centering
  \includegraphics[width=\columnwidth]{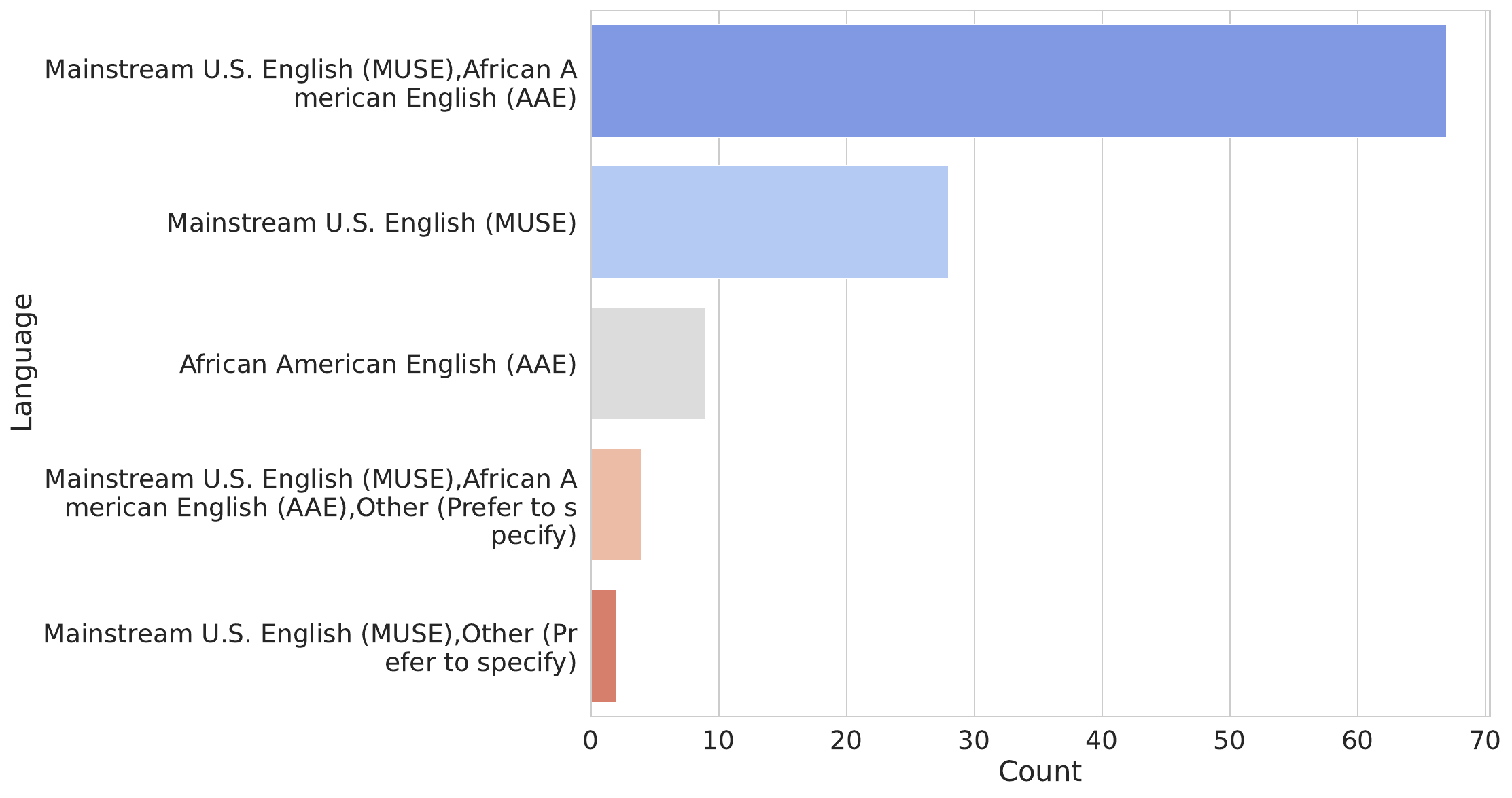} 
  \caption {\textit{Bar Plot of Language Proficiency Preferences}: This graph quantifies participant preferences for language proficiency in different varieties, focusing on Mainstream U.S. English (MUSE) and African American English (AAE). The bars represent the number of participants proficient in solely MUSE, solely AAE, a combination of both, and those with proficiencies that include other specified languages.}
  \label{fig:ling_div}
\end{figure*}

The ethnic group distribution showed significant representation from diverse backgrounds, and language proficiency varied widely among participants, which included proficiency in Mainstream U.S. English, African American English, and other specified languages. These factors underscore the multicultural and multilingual composition of our respondents. (see ~\autoref{fig:ling_div})



\begin{figure*}[t]
  \centering
  \includegraphics[width=\linewidth]{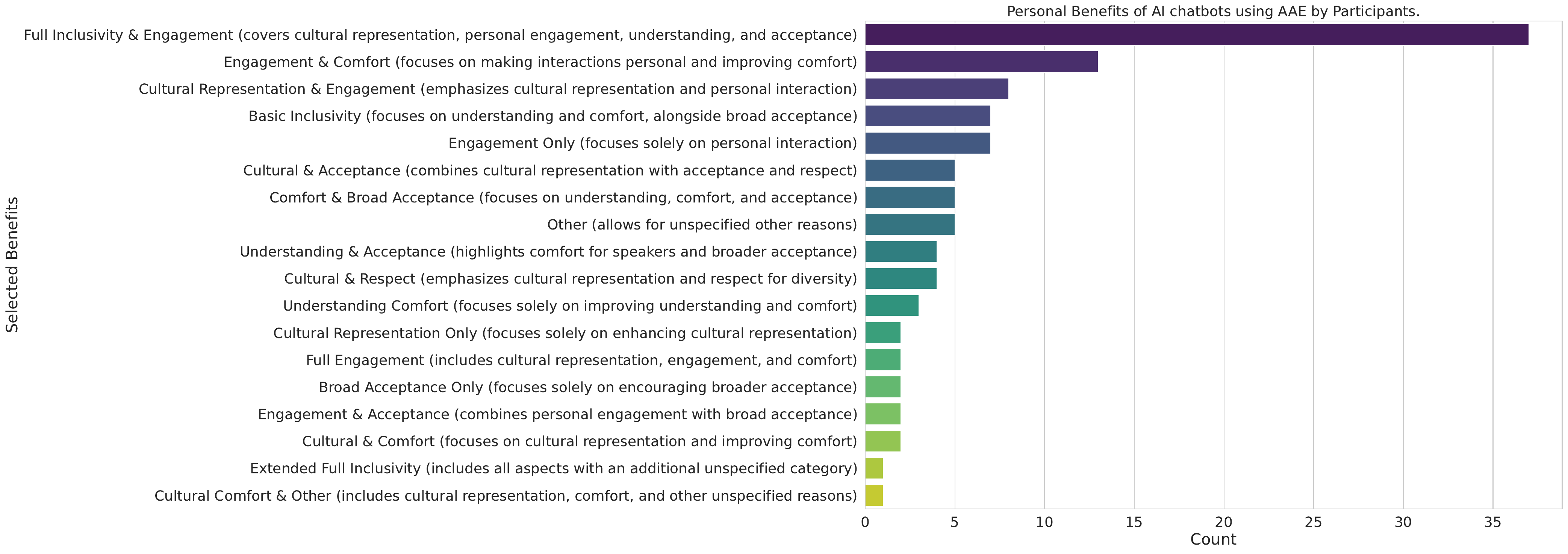} 
  \caption {\textit{Bar Plot of Perceived Benefits}: This graph illustrates the various benefits identified by participants when African American English (AAE) is incorporated into chatbot interactions. Each bar represents specific advantages such as enhanced cultural representation, personal engagement, and broader acceptance. 
  }
  \label{fig:benefits}
\end{figure*}

\begin{figure*}[t]
  \centering
  \includegraphics[width=\linewidth]{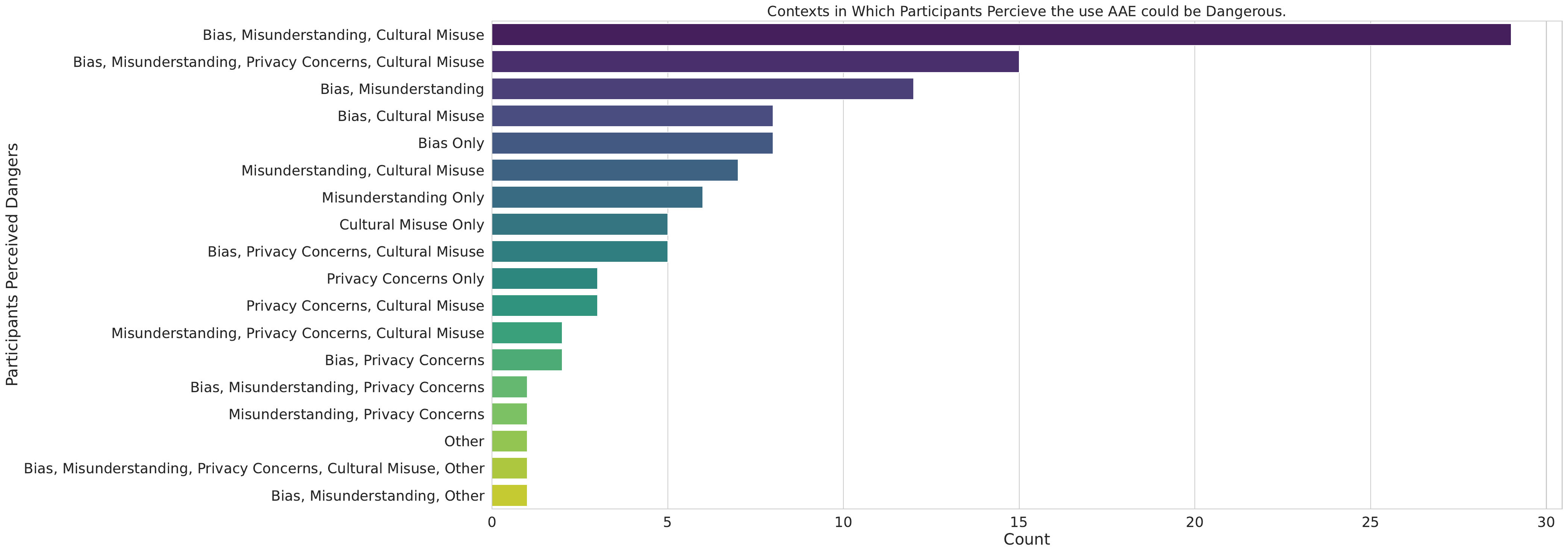} 
  \caption {\textit{Bar Plot of Participant Concerns}: This graph illustrates the range of selected concerns among participants regarding the integration of African American English (AAE) into chatbot technology. Each bar represents a distinct set of issues, from perpetuating stereotypes and biases to potential misunderstandings and fears of cultural appropriation. 
  }
  \label{fig:dangers}
\end{figure*}


\begin{figure*}[t]
  \centering
  \includegraphics[width=\linewidth]{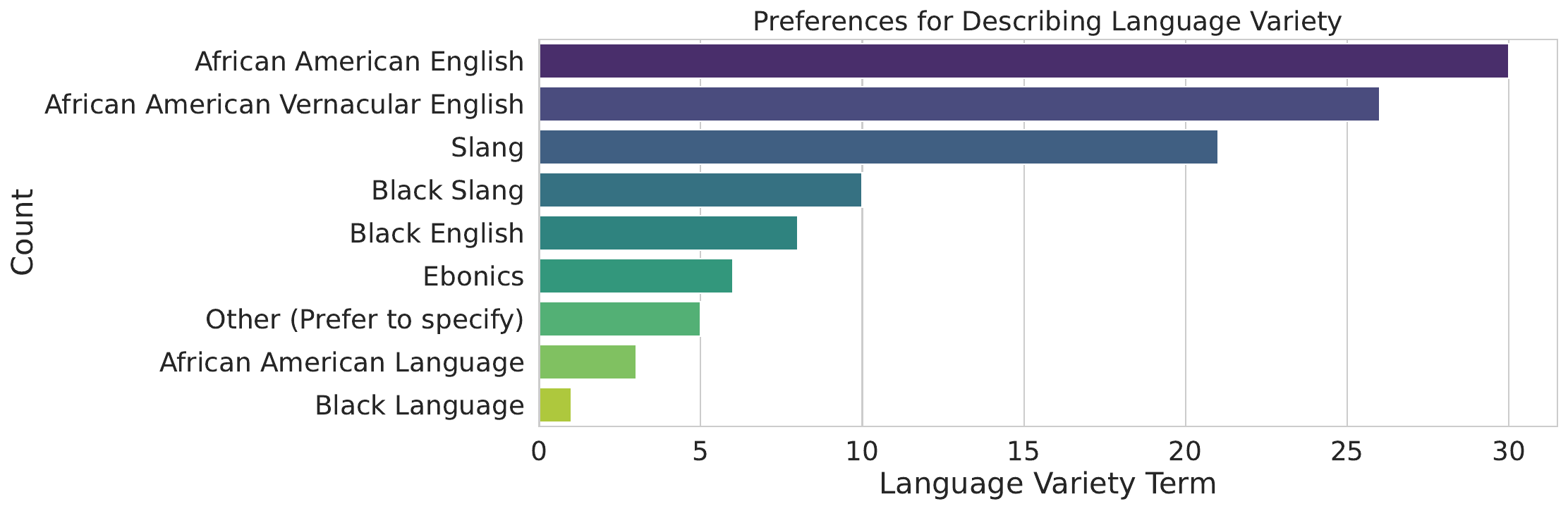} 
  \caption {\textit{Bar Plot of Terminology Preferences for AAE}: This graph presents the count of participants' preferences for various terms used to describe African American English. Each bar represents the popularity of terms such as `African American English', `African American Vernacular English', `Ebonics', and other variants. The plot underscores the diverse linguistic identities within the African American community and highlights the specific terminology that participants feel most accurately represents their language variety.}
  \label{fig:language variety term}
\end{figure*}

\begin{figure*}[t]
  \centering
  \includegraphics[width=\linewidth]{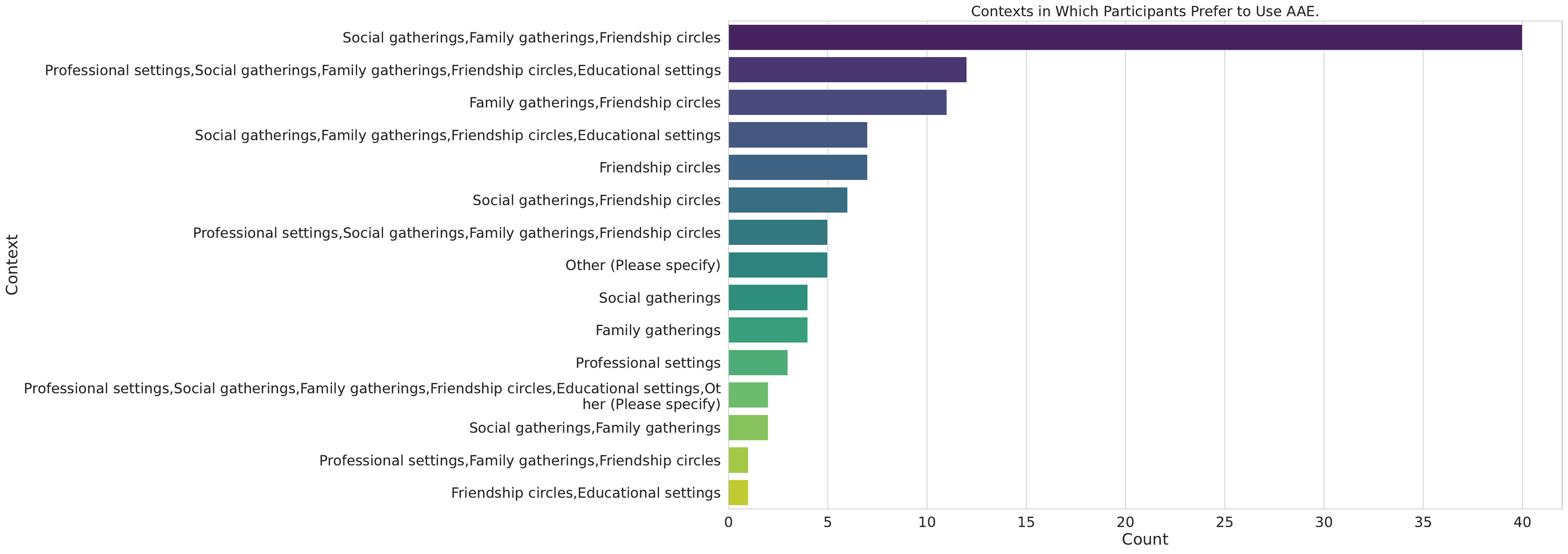} 
  \caption {\textit{Bar Plot of Contextual Preferences for Using AAE}: This graph displays the frequency of preferences among participants for using African American English (AAE) across various social and professional contexts. Each bar indicates the count of participants who prefer using AAE in settings ranging from personal interactions, such as family and friendship circles, to more formal environments like professional and educational settings. }
  \label{fig:lang_use_contx}
\end{figure*}

\begin{figure*}[t]
  \centering
  \includegraphics[width=\linewidth]{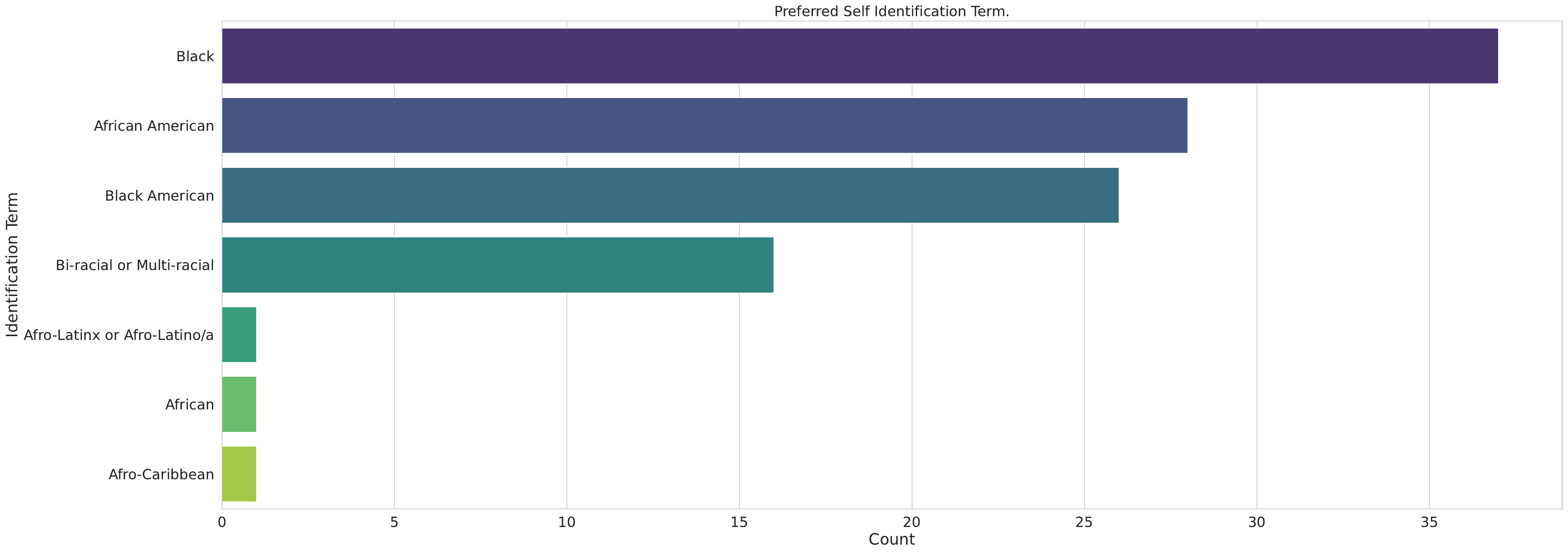} 
  \caption {\textit{Bar Plot of Preferred Self-Identification Terms}: This graph illustrates the distribution of preferred self-identification terms among respondents, highlighting the diversity within racial and ethnic identities. The terms range from `Black' and `African American' to more specific identities such as `Afro-Latinx' and `Afro-Caribbean.' Each bar represents the count of individuals who prefer each term, with `Black' and `African American' being the most common, followed by `Black American' and `Bi-racial or Multi-racial.' }
  \label{fig:identifying_term}
\end{figure*}


\end{document}